\documentclass{article}
\usepackage{typeset,times}

\usepackage{amsmath,amsfonts,bm}

\def\eqref#1{equation~\ref{#1}}
\def\1{\bm{1}}

\DeclareMathAlphabet{\mathsfit}{\encodingdefault}{\sfdefault}{m}{sl}
\SetMathAlphabet{\mathsfit}{bold}{\encodingdefault}{\sfdefault}{bx}{n}

\usepackage{hyperref}
\usepackage{url}

\usepackage{amsmath}
\usepackage{booktabs}
\usepackage{tabularx}
\usepackage{wrapfig}
\usepackage{caption}
\usepackage{graphicx}
\usepackage{xcolor}
\usepackage{listings}
\newcommand{\resMuseText}{64.0}
\newcommand{\resMuseAudio}{32.4}
\newcommand{\resMuseShared}{61.9}
\newcommand{\resMuseSelf}{58.6}

\newcommand{\resMuseFactorialHistory}{20.1}

\newcommand{\resMuseFactorialCurrent}{6.2}

\newcommand{\resMuseFactorialFull}{26.2}

\newcommand{\resMuseGap}{31.6}

\newcommand{\resMuseSelfRecovered}{883}
\newcommand{\resMuseSelfLost}{1171}
\newcommand{\resMuseSelfRecovery}{75.4}

\newcommand{\resMuseIRSurvival}{56.9}

\newcommand{\resMuseVESurvival}{68.0}

\newcommand{\resMuseIRJASurvival}{40.4}

\newcommand{\resMuseEnglishTTSDelta}{-0.3}

\newcommand{\resGProText}{63.0}
\newcommand{\resGProAudio}{54.4}

\newcommand{\resGProFactorialHistory}{5.7}

\newcommand{\resGProFactorialCurrent}{2.8}

\newcommand{\resGProFactorialFull}{6.4}

\newcommand{\resGProSelfRecovered}{347}
\newcommand{\resGProSelfLost}{493}
\newcommand{\resGProSelfRecovery}{70.4}

\newcommand{\resGProIMSurvival}{79.8}

\newcommand{\resGProVESurvival}{94.5}
\newcommand{\resGProIMZHSurvival}{72.6}

\newcommand{\resGProIMTHSurvival}{86.2}
\newcommand{\resGProIMVISurvival}{71.2}

\newcommand{\resGProEnglishTTSDelta}{0.6}

\newcommand{\resQNewText}{60.9}
\newcommand{\resQNewAudio}{60.2}

\newcommand{\resQNewIMZHSurvival}{93.0}

\newcommand{\resQNewIMVISurvival}{77.8}

\newcommand{\resDoubaoIMSurvival}{79.4}

\newcommand{\resDoubaoIRSurvival}{84.3}

\newcommand{\resDoubaoSCSurvival}{87.1}

\newcommand{\resDoubaoVESurvival}{91.9}

\newcommand{\resDoubaoIMDESurvival}{86.7}

\newcommand{\resDoubaoIMTHSurvival}{70.7}

\newcommand{\resQOldAudio}{38.1}

\newcommand{\resQOldEnglishTTSDeltaAbs}{4.5}

\newcommand{\resMuseGapCI}{[$+28.0$, $+35.1$]}

\newcommand{\resMuseHMinusCPoint}{$+13.6$}
\newcommand{\resMuseHMinusCCI}{[$+10.2$, $+17.1$]}

\newcommand{\resGProGapCI}{[$+6.3$, $+10.8$]}

\newcommand{\resGProHMinusCPoint}{$+3.0$}
\newcommand{\resGProHMinusCCI}{[$+1.0$, $+5.1$]}

\title{Do Multi-Turn Conversational Capabilities Carry Over from Text to Speech? A Parallel Multilingual Benchmark}

\author{
\begin{tabular}{c}
\textbf{Jiaming Zhou}$^{1,2}$\thanks{Work done during an internship at Alibaba Token Hub, Alibaba Group.}
\quad
\textbf{Pei Zhang}$^{2}$
\quad
\textbf{Shiwan Zhao}$^{1}$
\quad
\textbf{Zhijun Wang}$^{2}$ \\ 
\quad
\textbf{Baosong Yang}$^{2}$ 
\quad
\textbf{Yong Qin}$^{1}$\thanks{Yong Qin is corresponding author. Correspondence to: qinyong@nankai.edu.cn} \\[1ex]
\quad
\normalfont $^1$College of Computer Science, Nankai University
\quad
$^2$Alibaba Token Hub, Alibaba Group
\end{tabular}
}

\preprintfinalcopy
\begin{document}
\maketitle

\begin{abstract}
Successful multi-turn voice interaction requires models to retain information from earlier turns, follow persistent instructions, and track revisions to conversational state. Yet these abilities are often evaluated from transcripts, leaving open whether successes demonstrated from text carry over when the same user turns are presented as speech, and whether this transfer is consistent across languages. We introduce Omnilingua-AudioMC, a strictly parallel multilingual extension of Audio MultiChallenge, to study this question under controlled conditions. The benchmark contains 354 conversations and 1,404 rubric items across ten languages, with conversational content, turn structure, cross-turn dependencies, and evaluation criteria aligned across transcript and speech conditions. Across six speech-language models, the shift from reference transcripts to direct speech input is strongly model- and language-dependent: models with similar transcript performance can diverge substantially under speech, and their relative strengths across languages can change. Paired rubric-level analysis further shows that aggregate transcript--speech gaps can obscure which conversational requirements are actually preserved, because losses of transcript-demonstrated successes can be offset by gains on other items. We then diagnose recoverable speech degradation by selectively replacing preceding user turns, the current user turn, or both with transcripts generated by the evaluated model. For the two models with the largest transcript--speech gaps, replacing preceding user turns yields a significantly larger macro-level improvement than replacing only the current turn, and the same numerical ordering holds across all ten languages; in contrast, the strongest direct-speech model gains little from either intervention. These results show that transcript performance is an unreliable proxy for multilingual spoken conversational performance and, for models most affected by speech, identify the representation of preceding spoken context as an important source of recoverable degradation. Our code is publicly available at \url{https://github.com/QwenLM/Omnilingua-Bench/tree/main/Omnilingua-AudioMC}.
\end{abstract}

\section{Introduction}

Speech-language models are increasingly expected to sustain conversations across languages \citep{qwenteam2026qwen38omni,xu2025qwen3omni}. Successful interaction, however, requires more than understanding the current utterance. A response may depend on information introduced several turns earlier, instructions that remain active throughout the dialogue, or corrections that revise previously established conversational state. A model may therefore satisfy a conversational requirement when given a transcript of the interaction yet fail when the same user turns must be processed from speech. Understanding whether conversational successes demonstrated from transcripts carry over to speech is thus central to evaluating multilingual voice interaction.

Reference transcripts provide a controlled diagnostic baseline for this question. They preserve the wording of the spoken interaction---including hesitation, repetition, false starts, and self-repair---while removing the need to process audio. Comparing reference transcripts with speech therefore asks whether a model can satisfy the same conversational requirements when the user turns are represented in a different modality, rather than comparing different tasks or independently constructed examples.

To enable this comparison, we construct Omnilingua-AudioMC, a strictly parallel multilingual extension of Audio MultiChallenge (AudioMC) \citep{gosai2026audiomultichallenge}. The benchmark contains 354 conversations and 1,404 rubric items covering Inference Memory, Instruction Retention, Self Coherence, and Voice Editing. Starting from human-recorded English conversations, we construct aligned conditions in nine additional languages while preserving conversational content, turn structure, cross-turn dependencies, and evaluation criteria. English retains the original human recording, while the nine non-English versions are localized and synthesized; a synthetic-English condition provides a separate control for speech realization. Details of the multilingual construction and validation pipeline are provided in Section~\ref{sec:benchmark}.

\begin{wrapfigure}{R}{0.48\textwidth}
\vspace{-\intextsep}
\centering
\includegraphics[width=\linewidth]{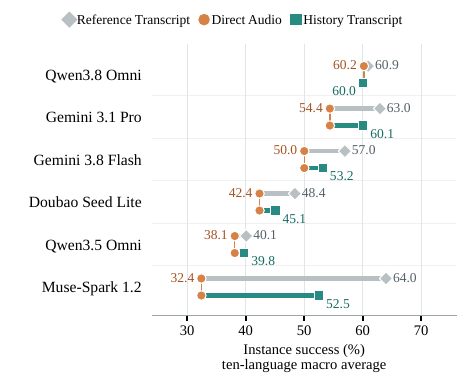}
\captionsetup{font=small,skip=5pt,justification=raggedright,singlelinecheck=false}
\caption{Overview of the six primary models: gray segments connect Reference Transcript and Direct Audio; teal points show History Transcript, which replaces preceding user turns with model-generated transcripts. Values are ten-language macro-mean instance success.}
\label{fig:intro_overview}
\end{wrapfigure}

Our first question is whether conversational performance demonstrated from reference transcripts carries over to direct speech input consistently across models and languages. The answer is clearly model-dependent. Among the six primary models, Muse-Spark has the highest average Reference Transcript instance success but suffers the largest drop under Direct Audio, whereas Qwen3.8 retains nearly the same aggregate performance and achieves the strongest Direct Audio results. The size of the shift also varies across languages, and relative language strengths can be reordered after moving from transcripts to speech (Figure~\ref{fig:intro_overview}). Reference-transcript performance is therefore not a reliable proxy for multilingual spoken conversational performance.

Aggregate scores alone, however, do not determine whether the same conversational requirements survive the modality change. On matched rubric items, some successes under Reference Transcript become failures under Direct Audio, while other items change in the opposite direction. These transitions can offset one another in the aggregate: even a model with a small overall transcript--speech gap need not preserve every reference-transcript success. We therefore complement marginal scores with paired rubric-level analysis that explicitly tracks which requirements remain satisfied under speech.

We next ask where the recoverable portion of this speech degradation arises. Using the same fixed-history evaluation protocol, we vary the representation of two parts of the user input: preceding user turns and the current user turn. This yields four intervention conditions. In \textit{Direct Audio}, both preceding and current user turns are presented as speech. In \textit{History Transcript}, only preceding user turns are replaced by transcripts generated by the evaluated model, while the current user turn remains audio. \textit{Current Transcript} makes the opposite substitution, retaining preceding user turns as audio and replacing only the current user turn with its model-generated transcript. \textit{Full Transcript} replaces both preceding and current user turns with model-generated transcripts. Historical assistant responses remain fixed as text throughout. These four conditions form a $2\times2$ diagnostic over history and current-turn representation, with Reference Transcript retained as a separate diagnostic baseline (Section~\ref{sec:experimental_setup}).

The intervention results reveal a strong model-dependent contrast. Muse-Spark and Gemini 3.1 Pro have the two largest Reference-Transcript--Direct-Audio gaps, and for both models History Transcript produces a larger macro-level improvement than Current Transcript, with the History--Current difference supported by conversation-level bootstrap intervals. The same numerical ordering holds across all ten evaluated languages. Qwen3.8 provides a different regime: its Direct Audio performance already remains close to its Reference Transcript performance, and neither history nor current-turn transcription produces a meaningful average improvement. The remaining models exhibit smaller and more language-dependent effects. These results show that transcript substitution is not intrinsically beneficial; rather, its benefit depends on where changing the representation of spoken input can recover performance.

For the models most affected by speech, these behavioral interventions point to preceding spoken context as an important source of recoverable degradation. They do not, however, identify a particular internal memory mechanism. History Transcript replaces several preceding user turns whereas Current Transcript replaces only the current turn, so the comparison does not isolate the position of the information from the amount of audio being replaced. The joint Full Transcript condition provides an additional diagnostic of how history- and current-turn-associated effects combine, but the corresponding interaction estimates remain secondary to the main localization result.

Our contributions are threefold:

\begin{itemize}

\item We introduce Omnilingua-AudioMC, a strictly parallel multilingual benchmark for persistent multi-turn conversational capabilities, containing 354 conversations and 1,404 rubric items across ten languages, with conversational content, cross-turn dependencies, and evaluation criteria aligned across transcript and speech conditions.

\item We show that transcript-to-speech transfer varies substantially across models and languages, and use paired rubric-level analysis to demonstrate that aggregate score gaps can obscure which conversational requirements are actually preserved under speech.

\item We introduce a controlled $2\times2$ intervention over preceding user history and the current user turn. For the two models with the largest transcript--speech gaps, replacing preceding user turns with model-generated transcripts yields a larger macro-level improvement than replacing only the current turn, while a strong Direct Audio model gains little from either intervention.

\end{itemize}

\section{Benchmark and Evaluation Framework}
    \label{sec:benchmark}

    \subsection{Benchmark Overview}
    \label{sec:benchmark_overview}

    Our benchmark tests whether persistent conversational capabilities demonstrated under \textit{Reference Transcript} remain available when the same content is presented as speech, building on Audio MultiChallenge (AudioMC; \citealp{gosai2026audiomultichallenge}). We retain four capabilities: \textit{Inference Memory} (IM), \textit{Instruction Retention} (IR), \textit{Self Coherence} (SC), and \textit{Voice Editing} (VE). These capabilities require a model to preserve or update information, instructions, and conversational states established in earlier turns.

    \begin{wraptable}{R}{0.47\textwidth}
    \vspace{-\intextsep}
    \captionsetup{font=small,skip=5pt,justification=raggedright,singlelinecheck=false}
    \caption{Benchmark composition. Dataset sizes are reported per aligned condition.}
    \label{tab:benchmark_statistics}
    \small
    \setlength{\tabcolsep}{3pt}
    \renewcommand{\arraystretch}{1.12}

    \begin{tabularx}{\linewidth}{@{}Xr@{}}
    \toprule
    \multicolumn{2}{@{}l}{\textbf{Dataset and conditions}} \\
    Conversations & 354 \\
    User turns & 1,827 \\
    Rubric items & 1,404 \\
    Reference Transcript / speech conditions & 10 / 11 \\
    \midrule
    \multicolumn{2}{@{}l}{\textbf{Conversations by capability}} \\
    Inference Memory (IM) & 82 \\
    Instruction Retention (IR) & 95 \\
    Self Coherence (SC) & 77 \\
    Voice Editing (VE) & 100 \\
    \midrule
    \multicolumn{2}{@{}l}{\textbf{Languages (10)}} \\
    \multicolumn{2}{@{}p{\linewidth}@{}}{
    \mbox{English (en)}, \mbox{Chinese (zh)}, \mbox{Japanese (ja)},
    \mbox{Spanish (es)}, \mbox{French (fr)}, \mbox{Russian (ru)},
    \mbox{German (de)}, \mbox{Italian (it)}, \mbox{Thai (th)},
    \mbox{Vietnamese (vi)}
    } \\
    \bottomrule
    \end{tabularx}
    \vspace{-6pt}
    \end{wraptable}

    The central design principle is \textbf{strict parallelism}. Each benchmark item is instantiated across languages and modalities while preserving the same conversation, turn structure, cross-turn dependencies, and evaluation criteria. This alignment is essential for attributing differences across conditions to language and input representation rather than to changes in conversational content or task requirements. The benchmark covers English and nine non-English languages (Table~\ref{tab:benchmark_statistics}), and headline analyses average over these ten languages. Each language has an aligned Reference Transcript condition and a speech condition. For English, the speech condition is the original human recording; we additionally synthesize a controlled English TTS speech condition (EN-TTS) that shares the English transcripts, isolating synthesis effects from language effects and reported separately (Section~\ref{sec:cross_modal_results}).

    For each conversation, the Reference Transcript and speech conditions form a matched pair evaluated with aligned rubric items: at the designated turn, user turns are presented as reference transcripts or speech with preceding assistant turns as fixed text, and the model generates only the designated response. We report rubric-level success and conversation-level success, the latter requiring all associated rubric items to be satisfied.

    All conditions are built from the same final conversation set: if a conversation cannot be reliably realized in a required language, parallel filtering removes it from the corresponding conditions (Section~\ref{sec:quality_control}), keeping comparisons aligned at the conversation and rubric levels. The same filtering also removes the upstream benchmark's memory-capability conversations that embed background sounds, a non-speech acoustic-event design; this work considers speech only.

    \subsection{Speech-Grounded Parallel Multilingual Construction}
    \label{sec:benchmark_construction}

    \paragraph{Preserving interactional structure.}
    \label{sec:interaction_parsing}
    \label{sec:localization}

    Construction starts from AudioMC's human-recorded English conversations and transcripts \citep{gosai2026audiomultichallenge}. Rather than translating utterances independently, localization preserves task-relevant content, conversational state transitions, cross-turn dependencies, and evaluation requirements in natural target-language speech; each user turn is localized with access to the English source history and previously localized turns, maintaining consistent entities, references, terminology, and state updates.

    Source-transcript analysis identifies task-relevant spoken phenomena, including hesitation, repetition, false starts, and self-repair, whose interactional structure is retained during localization. Self-repair is particularly important because earlier content may be explicitly withdrawn or revised: retaining only the final value of ``5 sticks of butter, make that 10, actually stick with 5'' would remove the conversational reasoning required by the original interaction. The resulting \textit{spoken-form text} serves both as the Reference Transcript input and as the linguistic content for speech synthesis, anchoring the two modalities to the same intended interaction.

    \paragraph{Controlling speech realization.}
    \label{sec:speech_realization}

    A single multilingual TTS system realizes the localized user turns with a fixed speaker-to-voice mapping per conversation, reused across synthesized languages; this reduces variation in speaker assignment without assuming that multilingual voice presets correspond to acoustically identical speakers. A synthetic-English condition using the unchanged English reference transcript and the same pipeline provides a within-language comparison with the original recordings. Detailed synthesis settings, delivery guidance, and recovery procedures appear in Appendix~\ref{app:construction_details}.

    \paragraph{Validating content and maintaining parallelism.}
    \label{sec:quality_control}

    Validation preserves task-relevant content and cross-condition alignment. Text review checks semantic fidelity, naturalness, cross-turn entity consistency, rubric applicability, and the order of task-critical revisions. Audio review combines transcripts from multiple ASR systems with model-based semantic comparison against the spoken-form reference to identify omissions, additions, or meaning-changing substitutions; because recognition errors can resemble synthesis defects, ambiguous cases receive additional review rather than automatic rejection.

    Confirmed defects trigger resynthesis with the same reference text and voice assignment, followed by renewed validation. During construction, \textbf{parallel filtering} propagates conversation-level exclusions across aligned conditions, preserving a shared evaluation set, and subsequent repairs retain the same conversational content and evaluation criteria. Appendix~\ref{app:construction_details} provides the full localization, synthesis, validation, and repair procedures.

\section{Experiments and Analysis}
\label{sec:experiments}

\subsection{Experimental Setup}
\label{sec:experimental_setup}

\paragraph{Models and evaluation.}
We evaluate six primary models: Muse-Spark 1.2 \citep{meta2026musespark12}, Gemini 3.1 Pro \citep{google2026gemini31pro}, Gemini 3.8 Flash \citep{google2026gemini38flash}, Qwen3.8/3.5 Omni \citep{qwenteam2026qwen38omni,qwenteam2026qwen35omni}, and Doubao Seed Lite \citep{bytedance2026doubaoseed20lite}. Following AudioMC's fixed-context, binary-rubric protocol \citep{gosai2026audiomultichallenge}, o4-mini \citep{openai2025o3o4mini} judges the response generated at the designated turn; robustness to this judge choice is assessed in Appendix~\ref{app:judge_robustness}. \textbf{Rubric success} pools passed criteria over scored criteria, while \textbf{instance success} requires all criteria of a conversation to pass. Unless a table states otherwise, macro averages weight the ten languages equally, with English human recordings included; English TTS is a synthesis control reported separately, and paired-comparison averages that lack an English export are noted in their captions. Marginal scores use available judgments, whereas paired and factorial analyses use comparison-specific common items or conversations. Marginal tables report point estimates, while conversation-level bootstrap intervals accompany the headline contrasts (Appendix~\ref{app:bootstrap_ci}); aggregation details appear in Appendix~\ref{app:result_coverage}.

\paragraph{Core input conditions.}
\textit{Reference Transcript} supplies the disfluency-preserving spoken-form text and serves as a diagnostic baseline. The remaining four conditions form a $2\times2$ intervention over the representation of preceding user turns and the current user turn: \textit{Direct Audio} presents both as audio; \textit{History Transcript} replaces only preceding user turns with model-generated transcripts; \textit{Current Transcript} replaces only the current user turn; and \textit{Full Transcript} replaces both. In the three transcript interventions, each evaluated model generates the transcript substitutions once, and those frozen substitutions are reused across its corresponding conditions. The factorial design probes where transcript substitution can recover performance. Improvements under History Transcript indicate greater behavioral sensitivity to the representation of preceding user turns, while improvements under Current Transcript indicate sensitivity to the representation of the current query. Full Transcript examines how the two intervention effects combine. These interventions localize recoverable degradation at the level of input representation rather than identifying a particular internal processing mechanism. Additional transcript-source diagnostics, including shared-recognizer transcripts and reference-transcript history, are reported in Appendix~\ref{app:transcript_sources}. These procedures add processing and are not compute-matched to Direct Audio.

\paragraph{Additional open models.}
Qwen2.5-Omni \citep{xu2025qwen25omni}, Voxtral-Mini \citep{mistral2025voxtral}, Qwen3-Omni \citep{xu2025qwen3omni}, and Voxtral-Small are evaluated locally under Reference Transcript and Direct Audio with the same judge and rubrics. Their condition-specific retained sets differ, so they are reported separately without paired or transcript-intervention analyses (Appendix~\ref{app:open_models}).

\subsection{Overall Multilingual Direct-Audio Performance}
\label{sec:overall_evaluation}

Table~\ref{tab:performance_overview} reports Direct Audio instance success for every language and primary model; the Avg., Std., and Range columns summarize its variation across the ten languages, and capability-level panels appear in Appendix Table~\ref{tab:audio_complete}. Qwen3.8 Omni achieves the highest overall Direct Audio instance success in each of the ten languages, with mean \resQNewAudio\%, followed by Gemini 3.1 Pro at \resGProAudio\%; Muse-Spark reaches only \resMuseAudio\% (Table~\ref{tab:modality_summary}). These absolute scores locate weaknesses; the matched Reference Transcript comparisons below establish modality-associated losses. Table~\ref{tab:input_conditions} reports the Reference Transcript baseline and the four core intervention conditions.

% Generated by paper/tools/update_experiment_results.py from results/exp.md.
\begin{table}[htbp]
\centering
\caption{Direct Audio \textbf{overall instance success} (\%) by language and model. Avg., population Std., and Range (max$-$min, in points) summarize the ten languages; EN uses human speech, other languages TTS; bold marks the highest value per column.}
\label{tab:performance_overview}
\begingroup
\fontsize{8.0}{10.0}\selectfont
\setlength{\tabcolsep}{1pt}
\renewcommand{\arraystretch}{1.06}
\begin{tabular*}{\linewidth}{@{}l@{\hspace{5pt}}|@{\hspace{5pt}\extracolsep{\fill}}*{10}{r}@{\hspace{5pt}}|@{\hspace{5pt}\extracolsep{\fill}}rrr@{}}
\toprule
\textbf{Model} & \textbf{EN} & \textbf{ZH} & \textbf{JA} & \textbf{ES} & \textbf{FR} & \textbf{RU} & \textbf{DE} & \textbf{IT} & \textbf{TH} & \textbf{VI} & \textbf{Avg.} & \textbf{Std.} & \textbf{Range} \\
\midrule
Qwen3.8 Omni & \textbf{61.6} & \textbf{60.7} & \textbf{60.5} & \textbf{64.1} & \textbf{59.6} & \textbf{61.0} & \textbf{61.6} & \textbf{60.7} & \textbf{54.0} & \textbf{58.2} & \textbf{60.2} & 2.5 & 10.1 \\
Gemini 3.1 Pro & 60.2 & 50.8 & 53.1 & 57.6 & 54.8 & 56.5 & 53.7 & 55.4 & 50.3 & 51.4 & 54.4 & 3.0 & 9.9 \\
Gemini 3.8 Flash & 52.0 & 51.7 & 48.6 & 51.4 & 48.9 & 53.1 & 50.6 & 51.4 & 48.0 & 44.4 & 50.0 & 2.4 & 8.7 \\
Doubao Seed Lite & 47.7 & 46.3 & 42.1 & 40.4 & 41.2 & 45.2 & 41.5 & 44.6 & 38.7 & 35.9 & 42.4 & 3.4 & 11.8 \\
Qwen3.5 Omni & 42.9 & 39.8 & 38.7 & 38.4 & 37.3 & 38.4 & 37.6 & 38.4 & 36.2 & 33.6 & 38.1 & 2.3 & 9.3 \\
Muse-Spark 1.2 & 39.8 & 28.7 & 29.4 & 34.2 & 32.5 & 37.9 & 33.7 & 32.6 & 27.1 & 28.3 & 32.4 & 4.0 & 12.7 \\
\bottomrule
\end{tabular*}
\endgroup
\end{table}

\subsection{RQ1: How does conversational capability change from Reference Transcript to Direct Audio across languages?}
\label{sec:cross_modal_results}

% Generated by paper/tools/update_experiment_results.py from results/exp.md.
\begin{table}[t]
\centering
\caption{Reference Transcript to Direct Audio shift in instance success (\%) by language and model. Language cells give Direct Audio minus Reference Transcript ({\color{teal!55!black}positive} in teal, {\color{red!65!black}negative} in red); Ref./Aud./$\Delta$ give ten-language macro means.}
\label{tab:modality_summary}
\fontsize{7.5}{9.2}\selectfont
\setlength{\tabcolsep}{1.2pt}
\begin{tabular*}{\linewidth}{@{}l@{\hspace{3pt}\extracolsep{\fill}}*{10}{r}@{\hspace{8pt}}rrr@{}}
\toprule
 & \multicolumn{10}{c}{Direct Audio $-$ Reference Transcript by language} & \multicolumn{3}{c}{Macro summary} \\
\cmidrule(lr){2-11}\cmidrule(lr){12-14}
\textbf{Model} & \textbf{EN} & \textbf{ZH} & \textbf{JA} & \textbf{ES} & \textbf{FR} & \textbf{RU} & \textbf{DE} & \textbf{IT} & \textbf{TH} & \textbf{VI} & \textbf{Ref.} & \textbf{Aud.} & \textbf{$\Delta$} \\
\midrule
Qwen3.8 Omni & {\color{gray!55}\fontsize{6.0}{7.2}\selectfont($0.0$)} & {\color{red!65!black}\fontsize{6.0}{7.2}\selectfont($-1.7$)} & {\color{teal!55!black}\fontsize{6.0}{7.2}\selectfont($+1.6$)} & {\color{red!65!black}\fontsize{6.0}{7.2}\selectfont($-2.6$)} & {\color{gray!55}\fontsize{6.0}{7.2}\selectfont($0.0$)} & {\color{red!65!black}\fontsize{6.0}{7.2}\selectfont($-2.0$)} & {\color{red!65!black}\fontsize{6.0}{7.2}\selectfont($-2.8$)} & {\color{gray!55}\fontsize{6.0}{7.2}\selectfont($0.0$)} & {\color{teal!55!black}\fontsize{6.0}{7.2}\selectfont($+3.2$)} & {\color{red!65!black}\fontsize{6.0}{7.2}\selectfont($-3.1$)} & 60.9 & 60.2 & {\color{red!65!black}\fontsize{6.0}{7.2}\selectfont($-0.7$)} \\
Gemini 3.1 Pro & {\color{red!65!black}\fontsize{6.0}{7.2}\selectfont($-2.8$)} & {\color{red!65!black}\fontsize{6.0}{7.2}\selectfont($-14.7$)} & {\color{red!65!black}\fontsize{6.0}{7.2}\selectfont($-11.9$)} & {\color{red!65!black}\fontsize{6.0}{7.2}\selectfont($-6.5$)} & {\color{red!65!black}\fontsize{6.0}{7.2}\selectfont($-8.5$)} & {\color{red!65!black}\fontsize{6.0}{7.2}\selectfont($-4.0$)} & {\color{red!65!black}\fontsize{6.0}{7.2}\selectfont($-9.6$)} & {\color{red!65!black}\fontsize{6.0}{7.2}\selectfont($-7.9$)} & {\color{red!65!black}\fontsize{6.0}{7.2}\selectfont($-11.6$)} & {\color{red!65!black}\fontsize{6.0}{7.2}\selectfont($-8.2$)} & 63.0 & 54.4 & {\color{red!65!black}\fontsize{6.0}{7.2}\selectfont($-8.6$)} \\
Gemini 3.8 Flash & {\color{red!65!black}\fontsize{6.0}{7.2}\selectfont($-7.3$)} & {\color{red!65!black}\fontsize{6.0}{7.2}\selectfont($-7.1$)} & {\color{red!65!black}\fontsize{6.0}{7.2}\selectfont($-9.3$)} & {\color{red!65!black}\fontsize{6.0}{7.2}\selectfont($-3.1$)} & {\color{red!65!black}\fontsize{6.0}{7.2}\selectfont($-5.6$)} & {\color{red!65!black}\fontsize{6.0}{7.2}\selectfont($-3.4$)} & {\color{red!65!black}\fontsize{6.0}{7.2}\selectfont($-8.2$)} & {\color{red!65!black}\fontsize{6.0}{7.2}\selectfont($-5.4$)} & {\color{red!65!black}\fontsize{6.0}{7.2}\selectfont($-7.9$)} & {\color{red!65!black}\fontsize{6.0}{7.2}\selectfont($-12.1$)} & 57.0 & 50.0 & {\color{red!65!black}\fontsize{6.0}{7.2}\selectfont($-6.9$)} \\
Doubao Seed Lite & {\color{teal!55!black}\fontsize{6.0}{7.2}\selectfont($+0.8$)} & {\color{red!65!black}\fontsize{6.0}{7.2}\selectfont($-4.3$)} & {\color{red!65!black}\fontsize{6.0}{7.2}\selectfont($-8.5$)} & {\color{red!65!black}\fontsize{6.0}{7.2}\selectfont($-8.8$)} & {\color{red!65!black}\fontsize{6.0}{7.2}\selectfont($-5.7$)} & {\color{red!65!black}\fontsize{6.0}{7.2}\selectfont($-3.7$)} & {\color{red!65!black}\fontsize{6.0}{7.2}\selectfont($-5.7$)} & {\color{red!65!black}\fontsize{6.0}{7.2}\selectfont($-3.7$)} & {\color{red!65!black}\fontsize{6.0}{7.2}\selectfont($-6.5$)} & {\color{red!65!black}\fontsize{6.0}{7.2}\selectfont($-14.4$)} & 48.4 & 42.4 & {\color{red!65!black}\fontsize{6.0}{7.2}\selectfont($-6.1$)} \\
Qwen3.5 Omni & {\color{teal!55!black}\fontsize{6.0}{7.2}\selectfont($+0.5$)} & {\color{red!65!black}\fontsize{6.0}{7.2}\selectfont($-1.2$)} & {\color{red!65!black}\fontsize{6.0}{7.2}\selectfont($-2.3$)} & {\color{red!65!black}\fontsize{6.0}{7.2}\selectfont($-3.1$)} & {\color{red!65!black}\fontsize{6.0}{7.2}\selectfont($-3.1$)} & {\color{red!65!black}\fontsize{6.0}{7.2}\selectfont($-2.0$)} & {\color{red!65!black}\fontsize{6.0}{7.2}\selectfont($-1.7$)} & {\color{red!65!black}\fontsize{6.0}{7.2}\selectfont($-0.9$)} & {\color{teal!55!black}\fontsize{6.0}{7.2}\selectfont($+0.9$)} & {\color{red!65!black}\fontsize{6.0}{7.2}\selectfont($-6.8$)} & 40.1 & 38.1 & {\color{red!65!black}\fontsize{6.0}{7.2}\selectfont($-2.0$)} \\
Muse-Spark 1.2 & {\color{red!65!black}\fontsize{6.0}{7.2}\selectfont($-24.9$)} & {\color{red!65!black}\fontsize{6.0}{7.2}\selectfont($-33.7$)} & {\color{red!65!black}\fontsize{6.0}{7.2}\selectfont($-34.6$)} & {\color{red!65!black}\fontsize{6.0}{7.2}\selectfont($-29.6$)} & {\color{red!65!black}\fontsize{6.0}{7.2}\selectfont($-34.2$)} & {\color{red!65!black}\fontsize{6.0}{7.2}\selectfont($-29.6$)} & {\color{red!65!black}\fontsize{6.0}{7.2}\selectfont($-31.3$)} & {\color{red!65!black}\fontsize{6.0}{7.2}\selectfont($-31.0$)} & {\color{red!65!black}\fontsize{6.0}{7.2}\selectfont($-34.2$)} & {\color{red!65!black}\fontsize{6.0}{7.2}\selectfont($-32.6$)} & 64.0 & 32.4 & {\color{red!65!black}\fontsize{6.0}{7.2}\selectfont($-31.6$)} \\
\bottomrule
\end{tabular*}
\end{table}

\paragraph{The strongest primary model on reference transcripts becomes the weakest in speech.}
Table~\ref{tab:modality_summary} contrasts the two conditions for the six primary models. Muse-Spark has the highest instance-level Reference Transcript mean, \resMuseText\%, but falls to \resMuseAudio\% under Direct Audio, a decline of \resMuseGap\ percentage points. Gemini 3.1 Pro drops from \resGProText\% to \resGProAudio\%. Qwen3.8, in contrast, remains close to its Reference Transcript performance: \resQNewText\% versus \resQNewAudio\%. Its small gap accompanies high absolute speech performance, unlike Qwen3.5, whose smaller decline occurs at a much lower Direct Audio score of \resQOldAudio\%. The headline gaps are stable under conversation-level resampling: ten-language 95\% bootstrap intervals are \resMuseGapCI\ percentage points for Muse-Spark and \resGProGapCI\ for Gemini 3.1 Pro (Appendix~\ref{app:bootstrap_ci}). These comparisons separate a large modality-associated deficit from low performance shared by both input conditions.

\paragraph{Language strengths under reference transcripts do not reliably predict speech strengths.}
The shift is strongly language-dependent (Table~\ref{tab:modality_summary}): Muse-Spark loses 25--35 points in every language, Gemini 3.1 Pro ranges from 3 to 15 points, and Qwen3.8 stays within about 3 points throughout. The per-language ordering also changes rather than being a uniform penalty on an otherwise unchanged ranking: Chinese is Gemini 3.1 Pro's strongest non-English reference language yet among its weakest under Direct Audio, while other models reorder their profiles in different ways. Qwen3.8 supplies a counterexample to a universal decline: Direct Audio exceeds Reference Transcript in Japanese and Thai, although Thai remains its weakest Direct Audio language overall (Table~\ref{tab:modality_summary}).

\paragraph{English TTS provides a within-language realization control.}
Replacing human-recorded English with synthetic English changes Muse's instance success by \resMuseEnglishTTSDelta\ points and Gemini 3.1 Pro's by +\resGProEnglishTTSDelta\ points, small relative to their non-English gaps, whereas Qwen3.5 loses \resQOldEnglishTTSDeltaAbs\ points (Table~\ref{tab:english_control}). This control measures sensitivity to the English realization; it neither decomposes the reference--audio gap nor establishes synthesis effects in the target languages.

The four local open models provide a further check on the distinction between absolute performance and modality gaps; because they use condition-specific retained sets, they are reported without paired outcomes (Appendix~\ref{app:open_models}).

\subsection{RQ2: Do aggregate Reference-Transcript--Direct-Audio gaps reflect preservation of the same conversational capabilities?}
\label{sec:capability_results}

\begin{figure}[t]
\centering
\begin{minipage}{\linewidth}
\centering
\textbf{\footnotesize (a) Paired outcomes}\par
\vspace{2pt}
\includegraphics[width=\linewidth]{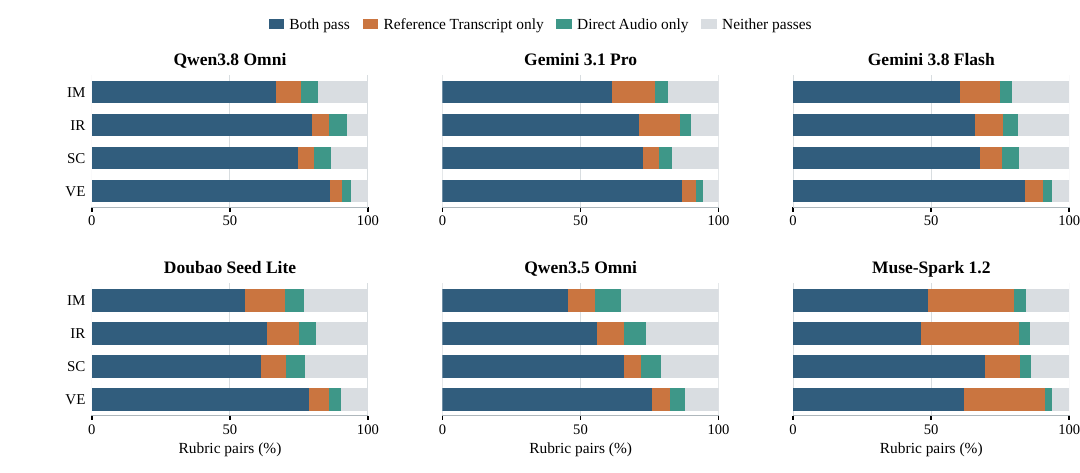}
\end{minipage}
\par\vspace{5pt}
\begin{minipage}{\linewidth}
\centering
\textbf{\footnotesize (b) Capability survival}\par
\vspace{2pt}
\includegraphics[width=\linewidth]{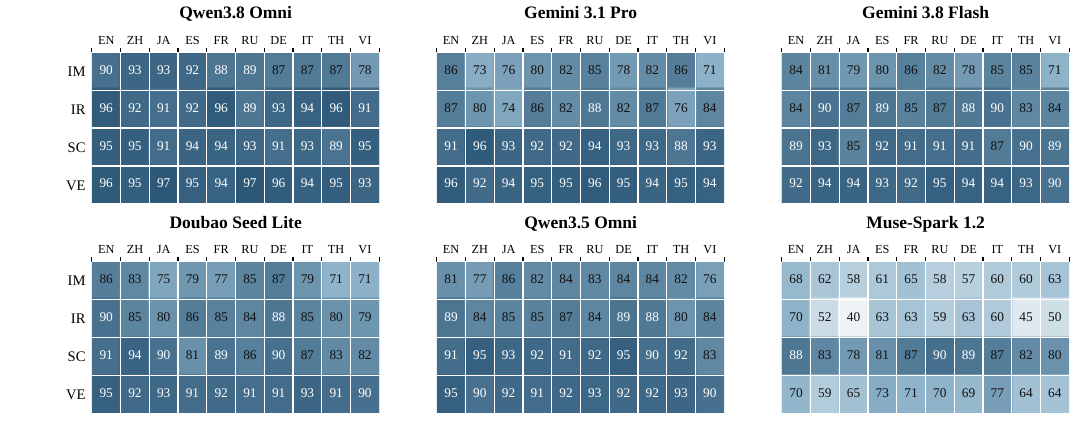}
\end{minipage}
\caption{\protect\input{tables/paired_capabilities_caption}}
\label{fig:paired_capabilities}
\end{figure}

Aggregate gaps can hide changes in which criteria are satisfied. Figure~\ref{fig:paired_capabilities} therefore separates paired rubric outcomes into success in both modalities, reference only, audio only, and neither. It also reports \textit{survival}: the fraction of reference-passing rubric items that remain successful under audio. These analyses use rubric-level pairs, whereas the overall scores discussed in RQ1 use instance success. The figure shows the six primary models in the same order as Table~\ref{tab:performance_overview}: Qwen3.8 Omni, Gemini 3.1 Pro, Gemini 3.8 Flash, Doubao Seed Lite, Qwen3.5 Omni, and Muse-Spark 1.2. This overview spans broad degradation, capability-specific vulnerabilities, and small net gaps. Full paired counts appear in Appendix~\ref{app:experimental_results}.

% Generated by paper/tools/update_experiment_results.py from results/exp.md.
\begin{table}[t]
\centering
\captionsetup{font=small,skip=5pt}
\caption{Survival of reference-passing rubric items under Direct Audio (\%), pooled over ten languages; language-level survival appears in Figure~\ref{fig:paired_capabilities}(b).}
\label{tab:survival_summary}
\small
\setlength{\tabcolsep}{2.5pt}
\renewcommand{\arraystretch}{1.12}
\begin{tabular}{@{}lrrrr@{}}
\toprule
\textbf{Model} & \textbf{IM} & \textbf{IR} & \textbf{SC} & \textbf{VE} \\
\midrule
Qwen3.8 Omni & 88.4 & 92.9 & 92.9 & 95.3 \\
Gemini 3.1 Pro & 79.8 & 82.7 & 92.6 & 94.5 \\
Gemini 3.8 Flash & 80.9 & 86.6 & 89.7 & 92.9 \\
Doubao Seed Lite & 79.4 & 84.3 & 87.1 & 91.9 \\
Qwen3.5 Omni & 81.9 & 85.3 & 91.4 & 92.1 \\
Muse-Spark 1.2 & 61.3 & 56.9 & 84.6 & 68.0 \\
\bottomrule
\end{tabular}
\end{table}

\paragraph{Memory and instruction retention are vulnerable, but the failure profile is broader for Muse.}
For Gemini 3.1 Pro, IM and IR show the largest reference-only shares among its capabilities, and only \resGProIMSurvival\% of its reference-passing IM criteria survive under audio, versus \resGProVESurvival\% for VE (Table~\ref{tab:survival_summary}); the vulnerability is language-specific, with IM survival at \resGProIMZHSurvival\% in Chinese and \resGProIMVISurvival\% in Vietnamese but \resGProIMTHSurvival\% in Thai (Figure~\ref{fig:paired_capabilities}b). Muse's losses extend beyond memory: IR survival is \resMuseIRSurvival\% and VE survival \resMuseVESurvival\%, with Japanese IR survival just \resMuseIRJASurvival\% (Table~\ref{tab:survival_summary}). A memory-only account would therefore miss substantial instruction and revision failures in this model.

Doubao likewise retains fewer reference-passing IM criteria (\resDoubaoIMSurvival\%) than of other capabilities (IR \resDoubaoIRSurvival\%, SC \resDoubaoSCSurvival\%, VE \resDoubaoVESurvival\%; Table~\ref{tab:survival_summary}), and its reference-only share exceeds its audio-only share by roughly 1.7 to 1, with uneven language-level IM survival from \resDoubaoIMDESurvival\% in German to \resDoubaoIMTHSurvival\% in Thai (Figure~\ref{fig:paired_capabilities}). Its moderate net decline therefore conceals more lost successes than the aggregate gap suggests; the lower absolute audio score also reflects criteria that fail in both modalities.

\paragraph{A small net gap does not imply preservation of the same successes.}
Qwen3.8's reference-only and audio-only shares are nearly balanced, at roughly 1.1 to 1, so opposing transitions largely cancel in its small aggregate gap, yet its IM survival falls to \resQNewIMVISurvival\% in Vietnamese against \resQNewIMZHSurvival\% in Chinese (Figure~\ref{fig:paired_capabilities}). Conversely, Gemini 3.1 Pro's relatively stable SC score coexists with nearly balanced shares. Paired outcomes thus identify failures that neither the overall gap nor the marginal capability scores can resolve.

\subsection{RQ3: Where does recoverable speech degradation arise: preceding history, the current turn, or both?}
\label{sec:history_results}

The four intervention conditions vary the representation of preceding user history and the current turn, with Reference Transcript a diagnostic baseline outside this factorial design (Section~\ref{sec:experimental_setup}). Table~\ref{tab:input_conditions} reports absolute ten-language macro-mean instance success, with each intervention's marginal contrast with Direct Audio in parentheses; these are marginal scores computed from available judgments, providing the descriptive condition-level comparison. For statistical comparisons, each contrast is re-estimated on its corresponding complete common conversation set, with conversation-level bootstrap intervals in Appendix~\ref{app:bootstrap_ci}.

% Generated by paper/tools/update_experiment_results.py from results/exp.md.
\begin{table}[htbp]
\centering
\caption{Ten-language macro-mean \textbf{instance success} (\%). Reference Transcript is a diagnostic baseline outside the $2\times2$ intervention. Parenthesized values give the marginal contrast with Direct Audio in points ({\color{teal!55!black}positive} in teal, {\color{red!65!black}negative} in red).}
\label{tab:input_conditions}
\begingroup
\fontsize{8.0}{10.0}\selectfont
\setlength{\tabcolsep}{1pt}
\renewcommand{\arraystretch}{1.06}
\begin{tabular*}{\linewidth}{@{}l@{\extracolsep{\fill}}r@{\hspace{16pt}}rrrr@{}}
\toprule
 & \multicolumn{1}{c}{Diagnostic baseline} & \multicolumn{4}{c}{$2\times2$ intervention} \\
\cmidrule(lr){2-2}\cmidrule(lr){3-6}
\textbf{Model} & \textbf{\shortstack{Reference\\Transcript}} & \textbf{\shortstack{Direct\\Audio}} & \textbf{\shortstack{History\\Transcript}} & \textbf{\shortstack{Current\\Transcript}} & \textbf{\shortstack{Full\\Transcript}} \\
\midrule
Qwen3.8 Omni & 60.9 & 60.2 & 60.1\,{\color{red!65!black}\footnotesize($-0.2$)} & 60.1\,{\color{red!65!black}\footnotesize($-0.1$)} & 58.8\,{\color{red!65!black}\footnotesize($-1.4$)} \\
Gemini 3.1 Pro & 63.0 & 54.4 & 60.1\,{\color{teal!55!black}\footnotesize($+5.7$)} & 57.1\,{\color{teal!55!black}\footnotesize($+2.8$)} & 60.8\,{\color{teal!55!black}\footnotesize($+6.4$)} \\
Gemini 3.8 Flash & 57.0 & 50.0 & 53.2\,{\color{teal!55!black}\footnotesize($+3.2$)} & 50.8\,{\color{teal!55!black}\footnotesize($+0.7$)} & 54.4\,{\color{teal!55!black}\footnotesize($+4.4$)} \\
Doubao Seed Lite & 48.4 & 42.4 & 45.1\,{\color{teal!55!black}\footnotesize($+2.7$)} & 43.4\,{\color{teal!55!black}\footnotesize($+1.1$)} & 46.3\,{\color{teal!55!black}\footnotesize($+3.9$)} \\
Qwen3.5 Omni & 40.1 & 38.1 & 39.8\,{\color{teal!55!black}\footnotesize($+1.6$)} & 38.4\,{\color{teal!55!black}\footnotesize($+0.2$)} & 38.1\,{\color{gray!55}\footnotesize($0.0$)} \\
Muse-Spark 1.2 & 64.0 & 32.4 & 52.5\,{\color{teal!55!black}\footnotesize($+20.1$)} & 38.6\,{\color{teal!55!black}\footnotesize($+6.2$)} & 58.6\,{\color{teal!55!black}\footnotesize($+26.2$)} \\
\bottomrule
\end{tabular*}
\endgroup
\end{table}

\begin{figure}[t]
\centering
\includegraphics[width=\linewidth]{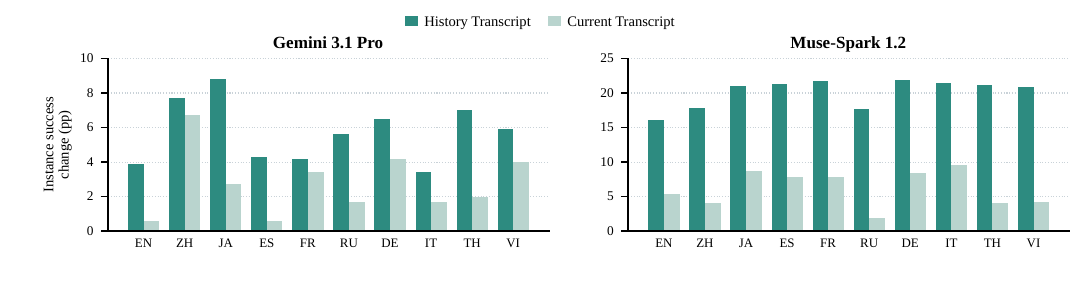}
\caption{History versus Current Transcript marginal contrasts (pp) relative to Direct Audio, per language, for the two models with the largest Direct-Audio deficits; bars above zero mark languages where the substitution raises instance success. Common-set bootstrap intervals appear in Appendix~\ref{app:bootstrap_ci}.}
\label{fig:intervention_languages}
\end{figure}

\paragraph{History replacement is most informative in the models with the largest Direct-Audio deficits.}
Muse gains \resMuseFactorialHistory\ points under History Transcript, compared with \resMuseFactorialCurrent\ under Current Transcript (Table~\ref{tab:input_conditions}); Gemini 3.1 Pro gains \resGProFactorialHistory\ and \resGProFactorialCurrent\ points, respectively. The History--Current difference itself is positive for both models: \resMuseHMinusCPoint\ points for Muse-Spark (95\% CI \resMuseHMinusCCI), with per-language intervals excluding zero in all ten languages, and \resGProHMinusCPoint\ points for Gemini 3.1 Pro (\resGProHMinusCCI). For both models, the History Transcript contrast is numerically larger than the Current Transcript contrast in every language (Figure~\ref{fig:intervention_languages}). History Transcript produces the larger improvement, indicating greater behavioral sensitivity to the representation of preceding spoken context; because it replaces several preceding turns while Current Transcript replaces only the current turn, the design does not separate history content from the amount of audio replaced. Full Transcript supplies the joint intervention, improving Muse by \resMuseFactorialFull\ points and Gemini 3.1 Pro by \resGProFactorialFull\ points.

\paragraph{The observed benefit is not universal.}
Qwen3.8, the strongest Direct Audio model, has near-zero average single-substitution contrasts and does not improve under Full Transcript; its language-level results are not uniformly zero, but this observational pattern shows that transcript substitution does not yield an average improvement for every model. The remaining models likewise do not support a universal history rule, showing language-dependent reversals in which substituting the current turn is the more beneficial intervention (Appendix Figure~\ref{fig:transcription_gains} and Table~\ref{tab:factorial_languages}).

\paragraph{The two substitutions recover partially overlapping sets of lost successes.}
Among conversations that succeed under Reference Transcript but fail under Direct Audio, History and Current Transcript each rescue cases that the other does not: across the six models, 50--70\% of recovered conversations are rescued by only one of the two substitutions, and History-only rescues exceed Current-only rescues in five of the six (Appendix~\ref{app:recovery_overlap}). This overlap evidence is separate from the aggregate-level question of additivity: the Full Transcript point estimate is close to the sum of the two single-substitution contrasts for several models, but the common-set interaction intervals span or nearly span zero, so departures from additivity are not established (Appendix~\ref{app:bootstrap_ci}).

\section{Related Work}

\paragraph{Parallel multilingual evaluation.}
XNLI and MGSM extend inference and mathematical reasoning across aligned examples \citep{conneau2018xnli,shi2023language}, and Multi-IF brings multilingual evaluation to multi-turn instruction following \citep{he2024multiif}. These text benchmarks motivate controlling task content across languages; ours additionally varies how the conversational context is presented, pairing reference transcripts with speech.

\paragraph{Spoken instructions and multimodal context.}
Speech-IFEval separates instruction-following assessment from speech perception \citep{lu2025speechifeval}. DOWIS provides parallel written and human-recorded spoken prompts across languages \citep{zufle2026dowis}, and MCIF evaluates crosslingual instructions over speech, text, and visual content from scientific talks \citep{papi2026mcif}. Our focus is complementary: preserving conversational information, constraints, and revisions across turns, and testing which input procedures recover failures on the same dialogue.

\paragraph{Multi-turn speech and transcript-mediated inference.}
AudioMC provides the human-recorded conversations and capability framework on which we build \citep{gosai2026audiomultichallenge}, comparing Direct Audio with a shared-ASR cascade with model-dependent gains and losses. We extend this comparison to strictly parallel multilingual conditions, track matched capability transitions, and localize recoverable degradation by varying the transcript representation of preceding history and the current turn. M3-DuplexBench examines multilingual full-duplex interaction \citep{fukuda2026m3duplexbench}; our fixed-context protocol instead targets content-dependent response correctness under controlled changes to input representation.

\section{Conclusion}

This work introduces Omnilingua-AudioMC, a strictly parallel multilingual benchmark for persistent conversational capabilities, grounded in human-recorded dialogue. The resulting comparisons show that reference-transcript performance is an unreliable proxy for spoken conversational ability: models can change their relative language strengths under speech, and small average modality gaps can conceal the loss of individual successes. Memory-related failures are prominent in several models, while instruction retention and revision tracking reveal additional vulnerabilities that an overall score would miss.
In the models with the largest Reference-Transcript--Direct-Audio gaps, History Transcript produces larger marginal recovery than Current Transcript in every language, with positive macro-level History--Current differences whose bootstrap intervals exclude zero. Qwen3.8's near-zero average interventions show that transcript substitution does not help every model, and other models contain language-level reversals in the preferred substitution. History and Current Transcript recover partially overlapping subsets of these lost successes, and the common-set interaction intervals span or nearly span zero, so departures from additivity are not established. These findings make history representation a key design consideration for multilingual voice interaction.

\subsection*{AI use statement}

In this work, we used generative AI tools to help implement and debug the evaluation and analysis code that produces the reported results; the code was reviewed and tested for correctness by the authors. We have not used generative AI tools for generating research ideas, experimental designs, or other tasks with required disclosure beyond the above, and the remaining required-disclosure categories are not applicable to this work. Additionally, we used generative AI tools to assist with surveying and organizing related literature; all cited works were verified against their original sources. We have reviewed all AI-assisted work. We take responsibility for the final content of this work, including code and artifacts produced with the aid of generative AI.

\subsection*{Ethics statement}

Our benchmark builds on the publicly released Audio MultiChallenge conversations and synthesizes all non-English speech with a commercial text-to-speech system; no new human-subject data collection is involved, and the source recordings are used under the terms of the original dataset. The English human/TTS control (Appendix~\ref{tab:english_control}) reports the sensitivity of this synthetic realization. The evaluation measures publicly available speech-enabled models in a fixed-context diagnostic protocol and does not involve personal data beyond the source conversations; we are not aware of harmful dual-use implications of the reported diagnostics.

\subsection*{Reproducibility statement}

The benchmark construction---localization, synthesis, validation, parallel filtering, and repair---is documented in Section~\ref{sec:benchmark} and Appendix~\ref{app:construction_details}, including synthesis settings and delivery guidance. The evaluation protocol, input conditions, and scoring conventions are specified in Section~\ref{sec:experimental_setup} and Appendix~\ref{app:result_coverage}, and the conversation-level bootstrap procedure appears in Appendix~\ref{app:bootstrap_ci}. All reported numbers derive from frozen model outputs and judge results over the same 354-conversation set.

\bibliography{references}
\bibliographystyle{typeset}

\clearpage
\appendix
% Appendix tables and figures are numbered per appendix section (Table A.1, B.1, ...).
\counterwithin{table}{section}
\counterwithin{figure}{section}
\counterwithin{lstlisting}{section}
% Generated by paper/tools/update_experiment_results.py from results/exp.md.
\section{Additional Experimental Results}

\label{app:experimental_results}

\subsection{Coverage and Comparison Scope}

\label{app:result_coverage}

All primary conditions target the same 354 conversations and 1,404 rubric items. A small number of judgments remained unavailable because of API failures and are not re-run. Marginal results use available judgments, whereas paired and factorial analyses use the corresponding comparison-specific common sets, so their denominators can differ from one another and from the marginal tables. Marginal tables report point estimates, and conversation-level bootstrap intervals accompany the headline contrasts (Appendix~\ref{app:bootstrap_ci}).

The evaluation uses AudioMC's fixed-context, binary-rubric protocol, but our rubric success pools criteria within each language. AudioMC's average rubric score first averages within each conversation; these aggregations differ when conversations have unequal numbers of criteria. Survival conditions on reference-passing paired rubric items. Transcription gains and recovery use comparison-specific common conversation sets. Recovery is the fraction of reference-success/audio-failure conversations that an intervention makes successful; it does not count new failures introduced elsewhere and should be read alongside net paired gain.

Historical assistant responses remain fixed text in every procedure: models generate only the designated response, rather than an online rollout of their own conversation histories.

Additional transcript-source diagnostics and implementation exceptions are described in Appendix~\ref{app:transcript_sources}.

Avg. is the ten-language macro mean (English human recordings plus nine non-English languages); the paired-comparison gains of Appendix~\ref{app:transcript_sources} lack an English export and their Avg. is the nine non-English languages, as noted in their captions. Std. is the population standard deviation and Range is the maximum minus minimum, both in percentage points. These dispersion measures use the reported language scores, rounded to one decimal place, and do not measure sampling uncertainty. A dash denotes unavailable results rather than zero.

% Generated by paper/tools/update_experiment_results.py from results/exp.md.
\begin{table}[p]
\centering
\caption{Direct-audio results in all ten languages. Each cell reports \textbf{rubric / instance} success (\%). EN uses human speech; other languages use synthesized speech. Bold denotes the best displayed value separately for each metric, language, and capability. Overall aggregates all items, not the four capability percentages. Some runs have incomplete scoring; see Appendix~\ref{app:result_coverage}.}
\label{tab:audio_complete}
\begingroup
\fontsize{8.2}{9.8}\selectfont
\setlength{\tabcolsep}{1pt}
\renewcommand{\arraystretch}{1.06}
\begin{tabular*}{\linewidth}{@{}l@{\hspace{3pt}\extracolsep{\fill}}*{10}{r}@{}}
\toprule
\multicolumn{11}{@{}l}{\textbf{(a) Inference Memory (IM)}} \\
\textbf{Model} & \textbf{EN} & \textbf{ZH} & \textbf{JA} & \textbf{ES} & \textbf{FR} & \textbf{RU} & \textbf{DE} & \textbf{IT} & \textbf{TH} & \textbf{VI} \\
\midrule
Qwen3.8 Omni & 72.5/58.5 & \textbf{74.5}/\textbf{61.0} & \textbf{75.2}/\textbf{61.0} & \textbf{79.1}/\textbf{67.1} & \textbf{73.9}/\textbf{61.0} & \textbf{73.9}/\textbf{61.0} & \textbf{73.2}/\textbf{62.2} & \textbf{73.2}/\textbf{61.0} & \textbf{69.3}/\textbf{53.7} & \textbf{66.0}/\textbf{51.2} \\
Gemini 3.1 Pro & \textbf{75.2}/\textbf{59.8} & 60.8/43.9 & 66.0/48.8 & 64.7/51.2 & 68.0/50.0 & 69.3/52.4 & 64.7/48.8 & 66.0/51.2 & 68.0/\textbf{53.7} & 60.1/43.9 \\
Gemini 3.8 Flash & 72.5/\textbf{59.8} & 64.1/47.6 & 64.1/47.6 & 62.7/43.9 & 63.4/46.3 & 68.6/56.1 & 66.0/50.0 & 64.7/51.2 & 65.4/46.3 & 57.5/39.0 \\
Doubao Seed Lite & 65.4/47.6 & 64.7/48.8 & 58.2/43.9 & 60.1/42.7 & 60.8/42.7 & 70.6/52.4 & 70.6/50.0 & 62.7/46.3 & 52.3/37.8 & 58.2/40.2 \\
Qwen3.5 Omni & 55.6/34.1 & 47.1/32.9 & 60.1/42.7 & 56.2/41.5 & 51.6/34.1 & 58.2/40.2 & 56.2/34.1 & 58.8/42.7 & 55.6/31.7 & 47.7/28.0 \\
Muse-Spark 1.2 & 62.7/41.5 & 54.2/32.9 & 50.3/31.2 & 51.6/31.7 & 55.6/32.9 & 52.9/32.9 & 50.3/29.3 & 52.3/32.9 & 52.9/30.5 & 51.0/32.9 \\
\addlinespace[3pt]
\midrule
\multicolumn{11}{@{}l}{\textbf{(b) Instruction Retention (IR)}} \\
\textbf{Model} & \textbf{EN} & \textbf{ZH} & \textbf{JA} & \textbf{ES} & \textbf{FR} & \textbf{RU} & \textbf{DE} & \textbf{IT} & \textbf{TH} & \textbf{VI} \\
\midrule
Qwen3.8 Omni & \textbf{89.9}/\textbf{80.0} & \textbf{86.5}/\textbf{69.6} & \textbf{83.8}/\textbf{66.3} & \textbf{86.0}/\textbf{75.8} & \textbf{89.2}/\textbf{78.9} & \textbf{83.8}/\textbf{67.4} & \textbf{87.4}/\textbf{75.8} & \textbf{87.4}/\textbf{75.8} & \textbf{83.1}/\textbf{65.3} & \textbf{85.6}/\textbf{71.6} \\
Gemini 3.1 Pro & 81.3/71.6 & 72.3/57.9 & 65.1/54.7 & 78.8/67.4 & 74.8/66.3 & 80.6/61.1 & 72.7/63.2 & 82.4/66.3 & 65.8/53.7 & 76.6/64.2 \\
Gemini 3.8 Flash & 75.9/54.7 & 71.6/54.7 & 66.2/48.4 & 73.4/61.1 & 69.8/52.6 & 72.3/53.7 & 73.0/51.6 & 77.0/57.9 & 64.7/46.3 & 70.1/48.4 \\
Doubao Seed Lite & 75.2/53.7 & 73.0/51.6 & 65.5/45.3 & 71.2/49.5 & 68.3/47.4 & 67.6/48.4 & 70.5/45.3 & 69.8/54.7 & 63.3/44.2 & 67.6/43.2 \\
Qwen3.5 Omni & 66.2/50.5 & 62.9/45.3 & 60.1/42.1 & 67.3/45.3 & 62.9/45.3 & 66.9/45.3 & 64.4/44.2 & 66.2/43.2 & 58.6/42.1 & 64.0/42.1 \\
Muse-Spark 1.2 & 66.9/53.7 & 43.2/24.2 & 36.1/22.3 & 56.5/43.2 & 55.4/36.8 & 56.1/43.2 & 55.4/40.0 & 54.7/36.8 & 37.8/21.1 & 43.5/27.4 \\
\addlinespace[3pt]
\midrule
\multicolumn{11}{@{}l}{\textbf{(c) Self Coherence (SC)}} \\
\textbf{Model} & \textbf{EN} & \textbf{ZH} & \textbf{JA} & \textbf{ES} & \textbf{FR} & \textbf{RU} & \textbf{DE} & \textbf{IT} & \textbf{TH} & \textbf{VI} \\
\midrule
Qwen3.8 Omni & \textbf{82.3}/\textbf{41.6} & \textbf{81.4}/\textbf{49.4} & 77.9/\textbf{50.6} & \textbf{83.3}/\textbf{48.1} & \textbf{80.4}/\textbf{40.3} & 80.4/45.5 & \textbf{81.1}/\textbf{48.1} & \textbf{80.4}/\textbf{42.9} & \textbf{77.9}/36.4 & \textbf{83.0}/\textbf{48.1} \\
Gemini 3.1 Pro & 76.0/40.3 & 79.8/42.9 & \textbf{81.1}/48.1 & 77.6/44.2 & 74.8/\textbf{40.3} & \textbf{81.7}/46.8 & 77.0/37.7 & 77.9/40.3 & 70.7/32.5 & 76.0/36.4 \\
Gemini 3.8 Flash & 72.9/32.5 & 77.0/40.3 & 71.0/40.3 & 76.7/39.0 & 75.1/36.4 & 77.3/40.3 & 76.3/41.6 & 70.3/33.8 & 72.2/40.3 & 72.9/33.8 \\
Doubao Seed Lite & 72.2/29.9 & 73.5/37.7 & 71.6/36.4 & 64.0/26.0 & 67.2/26.0 & 69.4/31.2 & 66.2/27.3 & 67.8/24.7 & 64.0/28.6 & 66.6/26.0 \\
Qwen3.5 Omni & 77.0/37.7 & 78.2/40.3 & 71.0/35.1 & 72.6/31.2 & 73.5/35.1 & 73.5/28.6 & 74.8/37.7 & 69.7/29.9 & 70.3/32.5 & 69.7/32.5 \\
Muse-Spark 1.2 & 76.3/40.3 & 73.5/39.0 & 67.2/42.9 & 70.3/37.7 & 74.1/35.1 & \textbf{81.7}/\textbf{50.6} & 78.2/39.0 & 75.4/40.3 & 69.4/\textbf{41.6} & 69.1/37.7 \\
\addlinespace[3pt]
\midrule
\multicolumn{11}{@{}l}{\textbf{(d) Voice Editing (VE)}} \\
\textbf{Model} & \textbf{EN} & \textbf{ZH} & \textbf{JA} & \textbf{ES} & \textbf{FR} & \textbf{RU} & \textbf{DE} & \textbf{IT} & \textbf{TH} & \textbf{VI} \\
\midrule
Qwen3.8 Omni & 90.9/62.0 & \textbf{89.6}/\textbf{61.0} & \textbf{90.4}/\textbf{62.0} & \textbf{90.4}/63.0 & 88.4/55.0 & \textbf{92.1}/\textbf{67.0} & 89.5/58.0 & 88.0/\textbf{60.0} & \textbf{88.3}/57.0 & 87.2/\textbf{59.0} \\
Gemini 3.1 Pro & \textbf{91.3}/\textbf{65.0} & 87.0/56.0 & 90.2/59.0 & 89.9/\textbf{64.0} & \textbf{90.1}/\textbf{59.0} & 90.1/63.0 & \textbf{89.9}/\textbf{61.0} & \textbf{89.3}/\textbf{60.0} & 87.8/\textbf{58.0} & \textbf{88.9}/57.0 \\
Gemini 3.8 Flash & 87.7/58.0 & 89.2/\textbf{61.0} & 88.6/56.0 & 87.8/58.0 & 86.1/57.0 & 88.4/60.0 & 87.2/57.0 & 87.5/59.0 & 85.5/57.0 & 84.6/53.0 \\
Doubao Seed Lite & 87.7/56.0 & 83.5/46.0 & 82.0/42.0 & 83.2/41.0 & 84.1/46.0 & 81.4/47.0 & 82.5/42.0 & 84.6/49.0 & 81.4/42.0 & 81.9/33.0 \\
Qwen3.5 Omni & 84.1/47.0 & 82.5/40.0 & 79.6/35.0 & 80.8/35.0 & 81.2/34.0 & 81.1/38.0 & 79.6/34.0 & 82.6/37.0 & 80.5/37.0 & 79.6/31.0 \\
Muse-Spark 1.2 & 67.1/25.0 & 56.2/21.4 & 61.7/24.2 & 69.8/25.0 & 67.4/26.0 & 67.2/27.0 & 65.0/27.3 & 71.8/22.2 & 60.2/19.0 & 60.1/18.2 \\
\addlinespace[3pt]
\midrule
\multicolumn{11}{@{}l}{\textbf{(e) Overall}} \\
\textbf{Model} & \textbf{EN} & \textbf{ZH} & \textbf{JA} & \textbf{ES} & \textbf{FR} & \textbf{RU} & \textbf{DE} & \textbf{IT} & \textbf{TH} & \textbf{VI} \\
\midrule
Qwen3.8 Omni & \textbf{86.8}/\textbf{61.6} & \textbf{85.5}/\textbf{60.7} & \textbf{84.6}/\textbf{60.5} & \textbf{86.7}/\textbf{64.1} & \textbf{85.2}/\textbf{59.6} & \textbf{85.8}/\textbf{61.0} & \textbf{85.4}/\textbf{61.6} & \textbf{84.5}/\textbf{60.7} & \textbf{82.8}/\textbf{54.0} & \textbf{83.6}/\textbf{58.2} \\
Gemini 3.1 Pro & 84.1/60.2 & 79.6/50.8 & 80.6/53.1 & 82.2/57.6 & 81.2/54.8 & 84.0/56.5 & 80.8/53.7 & 82.8/55.4 & 77.4/50.3 & 80.4/51.4 \\
Gemini 3.8 Flash & 80.3/52.0 & 80.2/51.7 & 77.5/48.6 & 79.7/51.4 & 77.9/48.9 & 80.6/53.1 & 79.6/50.6 & 79.1/51.4 & 76.2/48.0 & 76.1/44.4 \\
Doubao Seed Lite & 79.3/47.7 & 77.1/46.3 & 73.8/42.1 & 74.0/40.4 & 74.6/41.2 & 74.8/45.2 & 75.1/41.5 & 75.5/44.6 & 70.7/38.7 & 73.0/35.9 \\
Qwen3.5 Omni & 75.9/42.9 & 73.8/39.8 & 71.7/38.7 & 73.6/38.4 & 72.6/37.3 & 74.1/38.4 & 72.9/37.6 & 73.9/38.4 & 71.2/36.2 & 70.8/33.6 \\
Muse-Spark 1.2 & 68.7/39.8 & 57.3/28.7 & 56.6/29.4 & 65.3/34.2 & 65.2/32.5 & 66.7/37.9 & 64.5/33.7 & 67.0/32.6 & 57.1/27.1 & 57.9/28.3 \\
\bottomrule
\end{tabular*}
\endgroup
\end{table}

\subsection{Full Factorial Results}

\label{app:full_factorial_results}

\begin{figure}[htbp]

\centering

\includegraphics[width=\linewidth]{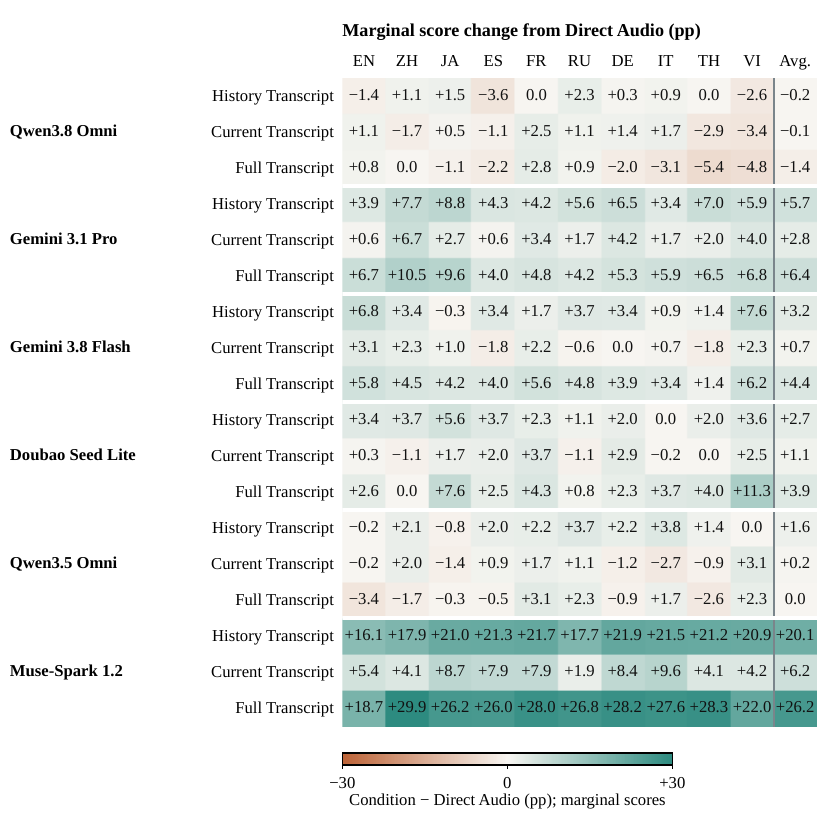}

\caption{Language-level marginal instance-success contrasts with Direct Audio (pp). All six models appear in the Table~\ref{tab:input_conditions} order, with three rows each: History Transcript, Current Transcript, and Full Transcript. EN is first; Avg. is the ten-language macro mean (English human recordings included). Contrasts use available judgments rather than a common four-condition set. The same contrasts appear numerically in Table~\ref{tab:factorial_languages}.}

\label{fig:transcription_gains}

\end{figure}

The full instance-level and rubric-level factorial results behind the marginal contrasts of Table~\ref{tab:input_conditions} appear in Tables~\ref{tab:instance_conditions} and~\ref{tab:rubric_conditions}; Appendix~\ref{app:factorial_details} details the four-condition comparison scope, and Figure~\ref{fig:transcription_gains} visualizes the language-level contrasts.

% Generated by paper/tools/update_experiment_results.py from results/exp.md.
\begin{table}[htbp]
\centering
\caption{Overall instance success (\%) by language and input condition. Avg. is the ten-language macro mean (English human recordings included). Entries use available judgments. Condition definitions and scoring exceptions appear in Appendix~\ref{app:result_coverage}.}
\label{tab:instance_conditions}
\begingroup
\fontsize{8.2}{9.1}\selectfont
\setlength{\tabcolsep}{1pt}
\renewcommand{\arraystretch}{1.06}
\begin{tabular*}{\linewidth}{@{}ll@{\extracolsep{\fill}}*{11}{r}@{}}
\toprule
\textbf{Model} & \textbf{Input} & \textbf{Avg.} & \textbf{EN} & \textbf{ZH} & \textbf{JA} & \textbf{ES} & \textbf{FR} & \textbf{RU} & \textbf{DE} & \textbf{IT} & \textbf{TH} & \textbf{VI} \\
\midrule
Qwen3.8 Omni & Reference Transcript & 60.9 & 61.6 & 62.4 & 58.9 & 66.7 & 59.6 & 63.0 & 64.4 & 60.7 & 50.8 & 61.3 \\
 & Direct Audio & 60.2 & 61.6 & 60.7 & 60.5 & 64.1 & 59.6 & 61.0 & 61.6 & 60.7 & 54.0 & 58.2 \\
 & History Transcript & 60.1 & 60.2 & 61.8 & 62.0 & 60.5 & 59.6 & 63.3 & 61.9 & 61.6 & 54.0 & 55.6 \\
 & Current Transcript & 60.1 & 62.7 & 59.0 & 61.0 & 63.0 & 62.1 & 62.1 & 63.0 & 62.4 & 51.1 & 54.8 \\
 & Full Transcript & 58.8 & 62.4 & 60.7 & 59.4 & 61.9 & 62.4 & 61.9 & 59.6 & 57.6 & 48.6 & 53.4 \\
\addlinespace[1pt]
\midrule
Gemini 3.1 Pro & Reference Transcript & 63.0 & 63.0 & 65.5 & 65.0 & 64.1 & 63.3 & 60.5 & 63.3 & 63.3 & 61.9 & 59.6 \\
 & Direct Audio & 54.4 & 60.2 & 50.8 & 53.1 & 57.6 & 54.8 & 56.5 & 53.7 & 55.4 & 50.3 & 51.4 \\
 & History Transcript & 60.1 & 64.1 & 58.5 & 61.9 & 61.9 & 59.0 & 62.1 & 60.2 & 58.8 & 57.3 & 57.3 \\
 & Current Transcript & 57.1 & 60.8 & 57.5 & 55.8 & 58.2 & 58.2 & 58.2 & 57.9 & 57.1 & 52.3 & 55.4 \\
 & Full Transcript & 60.8 & 66.9 & 61.3 & 62.7 & 61.6 & 59.6 & 60.7 & 59.0 & 61.3 & 56.8 & 58.2 \\
\addlinespace[1pt]
\midrule
Gemini 3.8 Flash & Reference Transcript & 57.0 & 59.3 & 58.8 & 57.9 & 54.5 & 54.5 & 56.5 & 58.8 & 56.8 & 55.9 & 56.5 \\
 & Direct Audio & 50.0 & 52.0 & 51.7 & 48.6 & 51.4 & 48.9 & 53.1 & 50.6 & 51.4 & 48.0 & 44.4 \\
 & History Transcript & 53.2 & 58.8 & 55.1 & 48.3 & 54.8 & 50.6 & 56.8 & 54.0 & 52.3 & 49.4 & 52.0 \\
 & Current Transcript & 50.8 & 55.1 & 54.0 & 49.6 & 49.6 & 51.1 & 52.5 & 50.6 & 52.1 & 46.2 & 46.7 \\
 & Full Transcript & 54.4 & 57.8 & 56.2 & 52.8 & 55.4 & 54.5 & 57.9 & 54.5 & 54.8 & 49.4 & 50.6 \\
\addlinespace[1pt]
\midrule
Doubao Seed Lite & Reference Transcript & 48.4 & 46.9 & 50.6 & 50.6 & 49.2 & 46.9 & 48.9 & 47.2 & 48.3 & 45.2 & 50.3 \\
 & Direct Audio & 42.4 & 47.7 & 46.3 & 42.1 & 40.4 & 41.2 & 45.2 & 41.5 & 44.6 & 38.7 & 35.9 \\
 & History Transcript & 45.1 & 51.1 & 50.0 & 47.7 & 44.1 & 43.5 & 46.3 & 43.5 & 44.6 & 40.7 & 39.5 \\
 & Current Transcript & 43.4 & 48.0 & 45.2 & 43.8 & 42.4 & 44.9 & 44.1 & 44.4 & 44.4 & 38.7 & 38.4 \\
 & Full Transcript & 46.3 & 50.3 & 46.3 & 49.7 & 42.9 & 45.5 & 46.0 & 43.8 & 48.3 & 42.7 & 47.2 \\
\addlinespace[1pt]
\midrule
Qwen3.5 Omni & Reference Transcript & 40.1 & 42.4 & 41.0 & 41.0 & 41.5 & 40.4 & 40.4 & 39.3 & 39.3 & 35.3 & 40.4 \\
 & Direct Audio & 38.1 & 42.9 & 39.8 & 38.7 & 38.4 & 37.3 & 38.4 & 37.6 & 38.4 & 36.2 & 33.6 \\
 & History Transcript & 39.8 & 42.7 & 41.9 & 37.9 & 40.4 & 39.5 & 42.1 & 39.8 & 42.2 & 37.6 & 33.6 \\
 & Current Transcript & 38.4 & 42.7 & 41.8 & 37.3 & 39.3 & 39.0 & 39.5 & 36.4 & 35.7 & 35.3 & 36.7 \\
 & Full Transcript & 38.1 & 39.5 & 38.1 & 38.4 & 37.9 & 40.4 & 40.7 & 36.7 & 40.1 & 33.6 & 35.9 \\
\addlinespace[1pt]
\midrule
Muse-Spark 1.2 & Reference Transcript & 64.0 & 64.7 & 62.4 & 64.0 & 63.8 & 66.7 & 67.5 & 65.0 & 63.6 & 61.3 & 60.9 \\
 & Direct Audio & 32.4 & 39.8 & 28.7 & 29.4 & 34.2 & 32.5 & 37.9 & 33.7 & 32.6 & 27.1 & 28.3 \\
 & History Transcript & 52.5 & 55.9 & 46.6 & 50.4 & 55.5 & 54.2 & 55.6 & 55.6 & 54.1 & 48.3 & 49.2 \\
 & Current Transcript & 38.6 & 45.2 & 32.8 & 38.1 & 42.1 & 40.4 & 39.8 & 42.1 & 42.2 & 31.2 & 32.5 \\
 & Full Transcript & 58.6 & 58.5 & 58.6 & 55.6 & 60.2 & 60.5 & 64.7 & 61.9 & 60.2 & 55.4 & 50.3 \\
\bottomrule
\end{tabular*}
\endgroup
\end{table}

% Generated by paper/tools/update_experiment_results.py from results/exp.md.
\begin{table}[htbp]
\centering
\caption{Overall rubric success (\%) by language and input condition. Avg. is the ten-language macro mean (English human recordings included). Entries use available judgments. Condition definitions and scoring exceptions appear in Appendix~\ref{app:result_coverage}.}
\label{tab:rubric_conditions}
\begingroup
\fontsize{8.2}{9.1}\selectfont
\setlength{\tabcolsep}{1pt}
\renewcommand{\arraystretch}{1.06}
\begin{tabular*}{\linewidth}{@{}ll@{\extracolsep{\fill}}*{11}{r}@{}}
\toprule
\textbf{Model} & \textbf{Input} & \textbf{Avg.} & \textbf{EN} & \textbf{ZH} & \textbf{JA} & \textbf{ES} & \textbf{FR} & \textbf{RU} & \textbf{DE} & \textbf{IT} & \textbf{TH} & \textbf{VI} \\
\midrule
Qwen3.8 Omni & Reference Transcript & 85.7 & 85.6 & 86.2 & 84.6 & 88.0 & 85.4 & 86.5 & 87.3 & 84.6 & 82.1 & 86.8 \\
 & Direct Audio & 85.1 & 86.8 & 85.5 & 84.6 & 86.7 & 85.2 & 85.8 & 85.4 & 84.5 & 82.8 & 83.6 \\
 & History Transcript & 84.7 & 85.2 & 85.8 & 84.2 & 85.8 & 84.3 & 85.6 & 85.2 & 85.1 & 82.1 & 83.3 \\
 & Current Transcript & 84.9 & 86.9 & 84.3 & 84.5 & 86.4 & 86.0 & 85.8 & 86.1 & 85.4 & 80.5 & 83.2 \\
 & Full Transcript & 84.5 & 85.9 & 85.2 & 85.1 & 86.4 & 86.5 & 85.6 & 85.2 & 83.0 & 78.6 & 83.6 \\
\addlinespace[1pt]
\midrule
Gemini 3.1 Pro & Reference Transcript & 86.1 & 87.1 & 87.1 & 87.0 & 87.1 & 86.1 & 85.1 & 86.7 & 86.2 & 83.6 & 85.0 \\
 & Direct Audio & 81.3 & 84.1 & 79.6 & 80.6 & 82.2 & 81.2 & 84.0 & 80.8 & 82.8 & 77.4 & 80.4 \\
 & History Transcript & 84.6 & 87.1 & 84.9 & 84.8 & 85.8 & 84.4 & 85.2 & 84.7 & 84.0 & 81.6 & 83.9 \\
 & Current Transcript & 83.3 & 85.2 & 82.6 & 83.1 & 83.8 & 83.9 & 84.1 & 83.9 & 82.8 & 79.8 & 83.3 \\
 & Full Transcript & 84.6 & 88.2 & 84.5 & 85.3 & 85.4 & 84.3 & 85.0 & 85.0 & 83.6 & 81.3 & 83.8 \\
\addlinespace[1pt]
\midrule
Gemini 3.8 Flash & Reference Transcript & 82.5 & 84.3 & 82.8 & 82.6 & 82.5 & 80.8 & 82.7 & 82.5 & 82.4 & 81.7 & 82.9 \\
 & Direct Audio & 78.7 & 80.3 & 80.2 & 77.5 & 79.7 & 77.9 & 80.6 & 79.6 & 79.1 & 76.2 & 76.1 \\
 & History Transcript & 80.5 & 83.0 & 80.4 & 78.1 & 81.5 & 79.0 & 81.8 & 81.1 & 80.8 & 78.1 & 81.2 \\
 & Current Transcript & 79.6 & 82.2 & 81.7 & 78.8 & 79.7 & 79.8 & 80.8 & 79.2 & 79.5 & 76.6 & 77.6 \\
 & Full Transcript & 81.4 & 83.3 & 82.4 & 80.8 & 83.0 & 81.4 & 83.4 & 81.6 & 80.1 & 77.6 & 79.9 \\
\addlinespace[1pt]
\midrule
Doubao Seed Lite & Reference Transcript & 78.5 & 78.7 & 79.0 & 78.6 & 80.3 & 77.6 & 78.4 & 77.4 & 78.1 & 76.7 & 80.1 \\
 & Direct Audio & 74.8 & 79.3 & 77.1 & 73.8 & 74.0 & 74.6 & 74.8 & 75.1 & 75.5 & 70.7 & 73.0 \\
 & History Transcript & 76.8 & 80.9 & 78.4 & 77.1 & 77.0 & 75.9 & 76.2 & 76.6 & 75.6 & 74.1 & 75.7 \\
 & Current Transcript & 75.6 & 79.2 & 76.6 & 75.0 & 76.1 & 75.9 & 75.3 & 76.7 & 76.7 & 72.4 & 72.1 \\
 & Full Transcript & 77.4 & 79.9 & 78.6 & 77.1 & 75.6 & 77.4 & 76.8 & 76.7 & 78.3 & 75.8 & 77.3 \\
\addlinespace[1pt]
\midrule
Qwen3.5 Omni & Reference Transcript & 73.7 & 76.0 & 75.2 & 71.2 & 74.8 & 73.7 & 73.7 & 73.5 & 73.4 & 70.2 & 75.3 \\
 & Direct Audio & 73.1 & 75.9 & 73.8 & 71.7 & 73.6 & 72.6 & 74.1 & 72.9 & 73.9 & 71.2 & 70.8 \\
 & History Transcript & 73.5 & 75.4 & 73.0 & 70.2 & 74.3 & 72.9 & 75.3 & 74.6 & 75.2 & 71.5 & 72.6 \\
 & Current Transcript & 73.1 & 75.1 & 74.0 & 72.2 & 72.6 & 74.1 & 74.7 & 73.1 & 71.6 & 69.3 & 74.6 \\
 & Full Transcript & 72.9 & 75.5 & 73.2 & 71.1 & 72.5 & 73.2 & 74.6 & 73.8 & 73.3 & 69.5 & 72.6 \\
\addlinespace[1pt]
\midrule
Muse-Spark 1.2 & Reference Transcript & 86.0 & 86.9 & 85.1 & 85.9 & 86.3 & 86.9 & 87.3 & 86.9 & 86.2 & 83.9 & 85.0 \\
 & Direct Audio & 62.6 & 68.7 & 57.3 & 56.6 & 65.3 & 65.2 & 66.7 & 64.5 & 67.0 & 57.1 & 57.9 \\
 & History Transcript & 79.9 & 82.2 & 77.5 & 77.7 & 82.5 & 78.9 & 81.6 & 81.6 & 81.6 & 77.3 & 78.4 \\
 & Current Transcript & 70.6 & 74.9 & 67.1 & 69.1 & 72.8 & 70.8 & 72.2 & 73.0 & 72.8 & 65.8 & 67.4 \\
 & Full Transcript & 83.2 & 83.0 & 83.5 & 82.2 & 84.6 & 82.5 & 85.4 & 85.0 & 84.5 & 79.8 & 81.2 \\
\bottomrule
\end{tabular*}
\endgroup
\end{table}

\subsubsection{History--Current Contrasts and Scoring Scope}

\label{app:factorial_details}

The two input factors are preceding user history and the current user turn. With A denoting audio and T the frozen model-generated transcript, (A,A) is Direct Audio, (T,A) is History Transcript, (A,T) is Current Transcript, and (T,T) is Full Transcript. Historical assistant responses remain fixed text. The text substitutions reuse the same frozen transcripts. Reference Transcript and reference-history inputs are separate controls, not cells of this factorial design.

All contrasts in Table~\ref{tab:factorial_effects} are computed from the marginal instance scores in Tables~\ref{tab:input_conditions} and~\ref{tab:instance_conditions}. History and Current are the two single-substitution contrasts relative to Direct Audio; Full substitutes both. The arithmetic interaction TT$-$TA$-$AT$+$AA equals Full$-$History$-$Current. Positive values indicate a larger full-substitution contrast than the sum of the two single substitutions; negative values indicate a smaller one. Neither sign establishes an internal processing mechanism.

Marginal condition scores and contrasts in Tables~\ref{tab:input_conditions} and~\ref{tab:instance_conditions} use available judgments. To separate scoring coverage from intervention effects, Appendix~\ref{app:bootstrap_ci} re-estimates each headline contrast on its corresponding complete common set; the factorial interaction is computed on the intersection complete across Direct Audio, History Transcript, Current Transcript, and Full Transcript. Its interval spans zero for five of the six models and lies below zero for Gemini 3.1 Pro ($-2.1$ points, 95\% CI $[-4.1, -0.03]$, whose upper endpoint rounds to zero at the table's integer precision), so interaction size and stability remain unestablished and additivity is a point-estimate observation. Current Transcript is reported at the Overall level in the marginal tables; the recovery-overlap analysis of Appendix~\ref{app:recovery_overlap} compares its conversation-level rescues with History Transcript on the four-arm common set.

% Generated by paper/tools/update_experiment_results.py from results/exp.md.
\begin{table}[htbp]
\centering
\caption{Transcript substitutions in the $2\times2$ design: ten-language macro-mean instance-score contrasts (pp). History, Current and Full subtract Direct Audio from History Transcript, Current Transcript and Full Transcript, respectively. Interaction is Full$-$History$-$Current. These are \textbf{marginal contrasts}, not common-set paired effects.}
\label{tab:factorial_effects}
\begingroup
\fontsize{8.5}{10.5}\selectfont
\setlength{\tabcolsep}{1pt}
\renewcommand{\arraystretch}{1.06}
\begin{tabular*}{0.76\linewidth}{@{}l@{\extracolsep{\fill}}rrrr@{}}
\toprule
\textbf{Model} & \textbf{History} & \textbf{Current} & \textbf{Full} & \textbf{Interaction} \\
\midrule
Qwen3.8 Omni & -0.2 & -0.1 & -1.4 & -1.2 \\
Gemini 3.1 Pro & +5.7 & +2.8 & +6.4 & -2.1 \\
Gemini 3.8 Flash & +3.2 & +0.7 & +4.4 & +0.4 \\
Doubao Seed Lite & +2.7 & +1.1 & +3.9 & +0.1 \\
Qwen3.5 Omni & +1.6 & +0.2 & 0.0 & -1.9 \\
Muse-Spark 1.2 & +20.1 & +6.2 & +26.2 & -0.2 \\
\bottomrule
\end{tabular*}
\endgroup
\end{table}

The main-text Table~\ref{tab:input_conditions} summarizes the average marginal contrasts; the tables below retain the per-language history/current contrasts and interaction details.

% Generated by paper/tools/update_experiment_results.py from results/exp.md.
\begin{table}[htbp]
\centering
\caption{Language-level history / current substitution contrasts in instance success (pp), each relative to marginal Direct Audio. Avg. is the ten-language macro mean (English human recordings included); individual language values use the reported rounded scores. These are descriptive contrasts on available judgments.}
\label{tab:factorial_languages}
\begingroup
\fontsize{6.6}{8.2}\selectfont
\setlength{\tabcolsep}{1pt}
\renewcommand{\arraystretch}{1.06}
\begin{tabular*}{\linewidth}{@{}l@{\extracolsep{\fill}}*{11}{r}@{}}
\toprule
\textbf{Model} & \textbf{Avg.} & \textbf{EN} & \textbf{ZH} & \textbf{JA} & \textbf{ES} & \textbf{FR} & \textbf{RU} & \textbf{DE} & \textbf{IT} & \textbf{TH} & \textbf{VI} \\
\midrule
Qwen3.8 Omni & -0.2/-0.1 & -1.4/+1.1 & +1.1/-1.7 & +1.5/+0.5 & -3.6/-1.1 & 0.0/+2.5 & +2.3/+1.1 & +0.3/+1.4 & +0.9/+1.7 & 0.0/-2.9 & -2.6/-3.4 \\
Gemini 3.1 Pro & +5.7/+2.8 & +3.9/+0.6 & +7.7/+6.7 & +8.8/+2.7 & +4.3/+0.6 & +4.2/+3.4 & +5.6/+1.7 & +6.5/+4.2 & +3.4/+1.7 & +7.0/+2.0 & +5.9/+4.0 \\
Gemini 3.8 Flash & +3.2/+0.7 & +6.8/+3.1 & +3.4/+2.3 & -0.3/+1.0 & +3.4/-1.8 & +1.7/+2.2 & +3.7/-0.6 & +3.4/0.0 & +0.9/+0.7 & +1.4/-1.8 & +7.6/+2.3 \\
Doubao Seed Lite & +2.7/+1.1 & +3.4/+0.3 & +3.7/-1.1 & +5.6/+1.7 & +3.7/+2.0 & +2.3/+3.7 & +1.1/-1.1 & +2.0/+2.9 & 0.0/-0.2 & +2.0/0.0 & +3.6/+2.5 \\
Qwen3.5 Omni & +1.6/+0.2 & -0.2/-0.2 & +2.1/+2.0 & -0.8/-1.4 & +2.0/+0.9 & +2.2/+1.7 & +3.7/+1.1 & +2.2/-1.2 & +3.8/-2.7 & +1.4/-0.9 & 0.0/+3.1 \\
Muse-Spark 1.2 & +20.1/+6.2 & +16.1/+5.4 & +17.9/+4.1 & +21.0/+8.7 & +21.3/+7.9 & +21.7/+7.9 & +17.7/+1.9 & +21.9/+8.4 & +21.5/+9.6 & +21.2/+4.1 & +20.9/+4.2 \\
\bottomrule
\end{tabular*}
\endgroup
\end{table}

\subsection{Conversation-Level Bootstrap Intervals}

\label{app:bootstrap_ci}

The headline contrasts are re-estimated by resampling source conversations rather than rubric items. A conversation counts as successful only when every rubric judged for it succeeds, matching the instance scores used elsewhere. Each model block draws one bag-of-source-conversations resample and applies the same sampled conversation identifiers to all ten language versions and all conditions, so the intervals exploit the parallel multilingual structure instead of treating the ten language versions of one source conversation as independent samples. Each contrast is computed on its own comparison-specific common set per language: the conditions involved in that contrast are jointly complete, and Reference Transcript therefore never restricts the factorial contrasts. The reported point estimates are these common-set estimates; the marginal estimates in Tables~\ref{tab:modality_summary} and~\ref{tab:input_conditions} use available judgments and can differ slightly.

The resulting intervals appear in Table~\ref{tab:bootstrap_ci}. Intervals use 10{,}000 replicates and report the 2.5th--97.5th percentiles; the 10-lang. column averages the per-language replicate contrasts with English human recordings included. These intervals measure benchmark-sample uncertainty over the observed conversations. They do not cover generation stochasticity; judge-family robustness is assessed in Appendix~\ref{app:judge_robustness}.

% Generated by paper/tools/update_experiment_results.py from results/exp.md.
\begin{table}[t]
\centering
\caption{Ninety-five percent conversation-level bootstrap intervals (pp) for six contrasts: Ref.$-$Direct, History, Current, and Full each subtract Direct Audio; H$-$C is History minus Current; Interaction is Full$-$History$-$Current$+$Direct. Each contrast uses its own comparison-specific common set per language (322 to 354 conversations per language); source conversations are resampled jointly across the ten language versions, so the macro column reflects source-level uncertainty. The macro cell gives the common-set point estimate with 2.5--97.5 percentiles of 10{,}000 replicates; intervals cover benchmark-sample uncertainty only.}
\label{tab:bootstrap_ci}
\fontsize{7.0}{8.6}\selectfont
\setlength{\tabcolsep}{1.2pt}
\begin{tabular*}{\linewidth}{@{}l@{\hspace{5pt}}l@{\hspace{3pt}\extracolsep{\fill}}*{10}{r}@{\hspace{8pt}}r@{}}
\toprule
 & & \multicolumn{10}{c}{Per language (95\% CI)} & Macro \\
\cmidrule(lr){3-12}\cmidrule(lr){13-13}
\textbf{Model} & \textbf{Contrast} & \textbf{EN} & \textbf{ZH} & \textbf{JA} & \textbf{ES} & \textbf{FR} & \textbf{RU} & \textbf{DE} & \textbf{IT} & \textbf{TH} & \textbf{VI} & \textbf{10-lang.} \\
\midrule
Qwen3.8 Omni & Ref.$-$Direct & $-5$--$5$ & $-3$--$6$ & $-6$--$3$ & $-2$--$7$ & $-5$--$5$ & $-3$--$7$ & $-2$--$8$ & $-5$--$5$ & $-8$--$2$ & $-2$--$8$ & $+0.8$ [$-1$, $2$] \\
 & History & $-6$--$3$ & $-3$--$5$ & $-3$--$6$ & $-8$--$0$ & $-5$--$5$ & $-3$--$7$ & $-5$--$5$ & $-3$--$5$ & $-5$--$5$ & $-7$--$2$ & $0$ [$-2$, $1$] \\
 & Current & $-3$--$5$ & $-7$--$3$ & $-4$--$5$ & $-5$--$3$ & $-2$--$7$ & $-4$--$6$ & $-3$--$6$ & $-3$--$6$ & $-8$--$2$ & $-8$--$1$ & $0$ [$-1$, $1$] \\
 & Full & $-4$--$5$ & $-5$--$5$ & $-5$--$3$ & $-7$--$3$ & $-2$--$7$ & $-4$--$6$ & $-7$--$3$ & $-8$--$1$ & $-10$--$-1$ & $-10$--$0$ & $-1.4$ [$-3$, $0$] \\
 & H$-$C & $-7$--$2$ & $-2$--$8$ & $-4$--$5$ & $-6$--$1$ & $-7$--$2$ & $-3$--$6$ & $-6$--$3$ & $-5$--$3$ & $-2$--$8$ & $-4$--$5$ & $0$ [$-2$, $1$] \\
 & Interaction & $-5$--$7$ & $-6$--$7$ & $-10$--$3$ & $-3$--$8$ & $-6$--$6$ & $-9$--$4$ & $-10$--$3$ & $-12$--$0$ & $-9$--$4$ & $-5$--$7$ & $-1.2$ [$-3$, $1$] \\
\midrule
Gemini 3.1 Pro & Ref.$-$Direct & $-2$--$7$ & $10$--$20$ & $7$--$17$ & $2$--$11$ & $4$--$13$ & $-1$--$9$ & $5$--$14$ & $3$--$12$ & $7$--$16$ & $3$--$13$ & $+8.6$ [$6$, $11$] \\
 & History & $0$--$8$ & $3$--$12$ & $4$--$14$ & $0$--$9$ & $-1$--$9$ & $1$--$10$ & $2$--$11$ & $-1$--$8$ & $3$--$12$ & $1$--$11$ & $+5.7$ [$4$, $8$] \\
 & Current & $-4$--$5$ & $3$--$11$ & $-2$--$7$ & $-3$--$5$ & $-1$--$8$ & $-3$--$6$ & $0$--$8$ & $-3$--$6$ & $-3$--$6$ & $-1$--$8$ & $+2.8$ [$1$, $4$] \\
 & Full & $3$--$11$ & $6$--$15$ & $5$--$15$ & $-1$--$8$ & $0$--$10$ & $0$--$8$ & $1$--$10$ & $1$--$11$ & $1$--$12$ & $1$--$12$ & $+6.4$ [$4$, $9$] \\
 & H$-$C & $-1$--$8$ & $-4$--$6$ & $1$--$11$ & $-1$--$8$ & $-3$--$5$ & $-1$--$9$ & $-2$--$7$ & $-3$--$6$ & $0$--$10$ & $-2$--$6$ & $+3.0$ [$1$, $5$] \\
 & Interaction & $-3$--$8$ & $-10$--$1$ & $-8$--$4$ & $-6$--$5$ & $-9$--$3$ & $-9$--$3$ & $-11$--$0$ & $-4$--$6$ & $-8$--$3$ & $-9$--$3$ & $-2.1$ [$-4$, $0$] \\
\midrule
Gemini 3.8 Flash & Ref.$-$Direct & $2$--$12$ & $3$--$12$ & $5$--$14$ & $-2$--$8$ & $1$--$10$ & $-1$--$8$ & $3$--$13$ & $1$--$10$ & $4$--$12$ & $7$--$17$ & $+6.9$ [$5$, $9$] \\
 & History & $3$--$11$ & $-1$--$8$ & $-5$--$5$ & $-1$--$8$ & $-3$--$6$ & $-1$--$8$ & $-2$--$8$ & $-4$--$5$ & $-3$--$6$ & $3$--$13$ & $+3.2$ [$1$, $5$] \\
 & Current & $-2$--$8$ & $-3$--$7$ & $-3$--$5$ & $-7$--$3$ & $-2$--$7$ & $-5$--$4$ & $-5$--$5$ & $-3$--$5$ & $-6$--$3$ & $-2$--$7$ & $+0.8$ [$-1$, $2$] \\
 & Full & $1$--$10$ & $0$--$9$ & $-1$--$9$ & $-1$--$9$ & $1$--$10$ & $0$--$10$ & $-1$--$9$ & $-1$--$8$ & $-3$--$6$ & $1$--$11$ & $+4.4$ [$2$, $7$] \\
 & H$-$C & $-1$--$8$ & $-3$--$6$ & $-6$--$3$ & $0$--$10$ & $-5$--$4$ & $-1$--$9$ & $-2$--$8$ & $-5$--$5$ & $-1$--$8$ & $1$--$10$ & $+2.4$ [$1$, $4$] \\
 & Interaction & $-10$--$2$ & $-7$--$5$ & $-3$--$9$ & $-3$--$9$ & $-5$--$8$ & $-4$--$8$ & $-5$--$6$ & $-4$--$7$ & $-4$--$8$ & $-10$--$2$ & $0$ [$-2$, $2$] \\
\midrule
Doubao Seed Lite & Ref.$-$Direct & $-6$--$4$ & $-1$--$9$ & $4$--$13$ & $4$--$14$ & $1$--$10$ & $-1$--$9$ & $1$--$10$ & $-1$--$8$ & $2$--$11$ & $9$--$20$ & $+6.0$ [$4$, $8$] \\
 & History & $-2$--$8$ & $-1$--$8$ & $1$--$10$ & $0$--$8$ & $-2$--$6$ & $-4$--$6$ & $-3$--$7$ & $-5$--$5$ & $-3$--$6$ & $-1$--$8$ & $+2.7$ [$1$, $4$] \\
 & Current & $-4$--$5$ & $-6$--$3$ & $-3$--$6$ & $-2$--$6$ & $-1$--$8$ & $-6$--$3$ & $-2$--$8$ & $-5$--$4$ & $-4$--$4$ & $-2$--$7$ & $+1.0$ [$0$, $2$] \\
 & Full & $-3$--$7$ & $-5$--$5$ & $3$--$12$ & $-2$--$7$ & $-1$--$9$ & $-4$--$6$ & $-3$--$7$ & $-1$--$9$ & $-1$--$8$ & $6$--$17$ & $+3.9$ [$2$, $6$] \\
 & H$-$C & $-1$--$7$ & $0$--$9$ & $-1$--$9$ & $-3$--$6$ & $-6$--$3$ & $-2$--$7$ & $-6$--$4$ & $-5$--$5$ & $-3$--$6$ & $-4$--$6$ & $+1.7$ [$0$, $3$] \\
 & Interaction & $-7$--$5$ & $-8$--$3$ & $-6$--$6$ & $-8$--$3$ & $-8$--$5$ & $-5$--$7$ & $-9$--$4$ & $-2$--$10$ & $-5$--$8$ & $-1$--$11$ & $0$ [$-2$, $2$] \\
\midrule
Qwen3.5 Omni & Ref.$-$Direct & $-5$--$4$ & $-3$--$6$ & $-2$--$7$ & $-1$--$7$ & $-1$--$7$ & $-3$--$6$ & $-3$--$6$ & $-3$--$5$ & $-5$--$4$ & $3$--$11$ & $+1.9$ [$0$, $4$] \\
 & History & $-5$--$4$ & $-3$--$7$ & $-5$--$3$ & $-2$--$6$ & $-2$--$6$ & $-1$--$8$ & $-2$--$7$ & $-1$--$8$ & $-3$--$6$ & $-5$--$4$ & $+1.6$ [$0$, $3$] \\
 & Current & $-5$--$4$ & $-2$--$6$ & $-6$--$3$ & $-3$--$5$ & $-2$--$6$ & $-3$--$6$ & $-5$--$3$ & $-7$--$2$ & $-5$--$3$ & $-1$--$8$ & $0$ [$-1$, $2$] \\
 & Full & $-8$--$1$ & $-6$--$3$ & $-5$--$5$ & $-5$--$4$ & $-1$--$8$ & $-3$--$7$ & $-5$--$3$ & $-3$--$6$ & $-7$--$2$ & $-2$--$7$ & $0$ [$-2$, $2$] \\
 & H$-$C & $-5$--$5$ & $-5$--$5$ & $-4$--$5$ & $-3$--$6$ & $-4$--$5$ & $-2$--$7$ & $-1$--$8$ & $2$--$11$ & $-2$--$6$ & $-8$--$1$ & $+1.4$ [$0$, $3$] \\
 & Interaction & $-9$--$3$ & $-12$--$0$ & $-4$--$7$ & $-9$--$3$ & $-6$--$5$ & $-9$--$4$ & $-8$--$4$ & $-5$--$7$ & $-9$--$3$ & $-7$--$6$ & $-1.9$ [$-4$, $0$] \\
\midrule
Muse-Spark 1.2 & Ref.$-$Direct & $19$--$31$ & $27$--$40$ & $28$--$40$ & $24$--$36$ & $29$--$40$ & $24$--$36$ & $26$--$37$ & $25$--$37$ & $29$--$40$ & $27$--$38$ & $+31.5$ [$28$, $35$] \\
 & History & $10$--$22$ & $12$--$24$ & $15$--$27$ & $15$--$27$ & $16$--$27$ & $12$--$23$ & $16$--$27$ & $16$--$28$ & $15$--$27$ & $15$--$26$ & $+20.1$ [$17$, $23$] \\
 & Current & $0$--$11$ & $-1$--$10$ & $4$--$13$ & $3$--$13$ & $3$--$14$ & $-4$--$8$ & $3$--$14$ & $3$--$15$ & $-1$--$9$ & $-1$--$9$ & $+6.2$ [$4$, $9$] \\
 & Full & $12$--$25$ & $24$--$36$ & $20$--$32$ & $20$--$32$ & $22$--$34$ & $21$--$32$ & $22$--$34$ & $22$--$34$ & $22$--$34$ & $16$--$28$ & $+26.2$ [$23$, $29$] \\
 & H$-$C & $5$--$17$ & $7$--$20$ & $4$--$18$ & $7$--$19$ & $8$--$20$ & $10$--$22$ & $7$--$19$ & $6$--$19$ & $11$--$23$ & $10$--$22$ & $+13.6$ [$10$, $17$] \\
 & Interaction & $-10$--$4$ & $1$--$16$ & $-10$--$3$ & $-10$--$3$ & $-8$--$5$ & $0$--$14$ & $-8$--$6$ & $-10$--$4$ & $-3$--$9$ & $-10$--$4$ & $0$ [$-2$, $2$] \\
\bottomrule
\end{tabular*}
\end{table}

\subsection{Recovery-Set Overlap}

\label{app:recovery_overlap}

For each model, Table~\ref{tab:recovery_overlap} reports the conversation-level overlap between History and Current Transcript rescues among Reference-success/Direct-failure cases on the four-arm common set (Reference, Direct Audio, History, Current, complete per conversation and language). Exclusive is the share of recovered conversations rescued by only one substitution; Overlap is the share rescued by both.

% Generated from data/recovery_overlap.json (exp_plan Task 2).
\begin{table}[htbp]
\centering
\caption{Conversation-level overlap between History and Current Transcript rescues among Reference-success/Direct-failure conversations, pooled over the ten languages on the four-arm common set (Reference, Direct Audio, History, Current; complete per conversation and language). Exclusive is the share of recovered conversations (H only $+$ C only $+$ Both) rescued by exactly one substitution; Overlap is the share rescued by both.}
\label{tab:recovery_overlap}
\begingroup
\fontsize{8.5}{10.5}\selectfont
\setlength{\tabcolsep}{5pt}
\renewcommand{\arraystretch}{1.1}
\begin{tabular}{@{}lrrrrrrr@{}}
\toprule
\textbf{Model} & \textbf{Eligible} & \textbf{H only} & \textbf{C only} & \textbf{Both} & \textbf{Neither} & \textbf{Exclusive (\%)} & \textbf{Overlap (\%)} \\
\midrule
Qwen3.8 Omni & 374 & 60 & 69 & 128 & 117 & 50.2 & 49.8 \\
Gemini 3.1 Pro & 530 & 167 & 62 & 177 & 124 & 56.4 & 43.6 \\
Gemini 3.8 Flash & 513 & 129 & 82 & 142 & 160 & 59.8 & 40.2 \\
Doubao Seed Lite & 491 & 115 & 63 & 130 & 183 & 57.8 & 42.2 \\
Qwen3.5 Omni & 347 & 65 & 62 & 95 & 125 & 57.2 & 42.8 \\
Muse-Spark 1.2 & 1270 & 523 & 151 & 285 & 311 & 70.3 & 29.7 \\
\bottomrule
\end{tabular}
\endgroup
\end{table}

\subsection{Capability-Level Reference and Audio Results}

Reference-transcript and direct-audio rubric success by capability and language for the six primary models appears in Table~\ref{tab:capability_rubric}, complementing the capability panels of Table~\ref{tab:audio_complete}.

% Generated by paper/tools/update_experiment_results.py from results/exp.md.
\begin{table}[htbp]
\centering
\caption{Reference-transcript / direct-audio rubric success (\%) by language and capability. Both values in each cell are rubric-level marginal rates; they are not conditional survival rates.}
\label{tab:capability_rubric}
\begingroup
\fontsize{8.2}{9.8}\selectfont
\setlength{\tabcolsep}{1pt}
\renewcommand{\arraystretch}{1.06}
\begin{tabular*}{\linewidth}{@{}l@{\hspace{3pt}\extracolsep{\fill}}*{10}{r}@{}}
\toprule
\multicolumn{11}{@{}l}{\textbf{Inference Memory (IM)}} \\
\textbf{Model} & \textbf{EN} & \textbf{ZH} & \textbf{JA} & \textbf{ES} & \textbf{FR} & \textbf{RU} & \textbf{DE} & \textbf{IT} & \textbf{TH} & \textbf{VI} \\
\midrule
Qwen3.8 Omni & 73.9/72.5 & 75.2/74.5 & 75.2/75.2 & 77.1/79.1 & 72.5/73.9 & 79.7/73.9 & 79.1/73.2 & 75.8/73.2 & 71.2/69.3 & 76.5/66.0 \\
Gemini 3.1 Pro & 78.4/75.2 & 81.0/60.8 & 79.1/66.0 & 76.5/64.7 & 76.5/68.0 & 73.9/69.3 & 81.0/64.7 & 76.5/66.0 & 75.8/68.0 & 72.5/60.1 \\
Gemini 3.8 Flash & 79.1/72.5 & 72.5/64.1 & 79.1/64.1 & 73.2/62.7 & 68.0/63.4 & 77.8/68.6 & 79.7/66.0 & 71.9/64.7 & 73.2/65.4 & 73.9/57.5 \\
Doubao Seed Lite & 70.6/65.4 & 70.6/64.7 & 69.3/58.2 & 70.6/60.1 & 72.5/60.8 & 71.9/70.6 & 73.9/70.6 & 69.9/62.7 & 64.7/52.3 & 66.0/58.2 \\
Qwen3.5 Omni & 58.2/55.6 & 51.6/47.1 & 54.9/60.1 & 58.8/56.2 & 55.6/51.6 & 57.5/58.2 & 56.2/56.2 & 60.1/58.8 & 47.7/55.6 & 52.3/47.7 \\
Muse-Spark 1.2 & 80.4/62.7 & 76.5/54.2 & 81.7/50.3 & 80.4/51.6 & 83.0/55.6 & 79.7/52.9 & 84.3/50.3 & 81.7/52.3 & 77.1/52.9 & 75.8/51.0 \\
\addlinespace[3pt]
\midrule
\multicolumn{11}{@{}l}{\textbf{Instruction Retention (IR)}} \\
\textbf{Model} & \textbf{EN} & \textbf{ZH} & \textbf{JA} & \textbf{ES} & \textbf{FR} & \textbf{RU} & \textbf{DE} & \textbf{IT} & \textbf{TH} & \textbf{VI} \\
\midrule
Qwen3.8 Omni & 88.1/89.9 & 86.9/86.5 & 85.9/83.8 & 88.8/86.0 & 82.7/89.2 & 87.8/83.8 & 86.7/87.4 & 87.4/87.4 & 76.3/83.1 & 87.1/85.6 \\
Gemini 3.1 Pro & 91.0/81.3 & 86.0/72.3 & 84.2/65.1 & 87.4/78.8 & 88.1/74.8 & 84.2/80.6 & 83.5/72.7 & 88.1/82.4 & 82.7/65.8 & 86.3/76.6 \\
Gemini 3.8 Flash & 80.6/75.9 & 72.7/71.6 & 73.0/66.2 & 75.9/73.4 & 75.5/69.8 & 73.4/72.3 & 75.2/73.0 & 79.5/77.0 & 75.5/64.7 & 78.8/70.1 \\
Doubao Seed Lite & 78.1/75.2 & 79.1/73.0 & 73.4/65.5 & 78.1/71.2 & 72.3/68.3 & 73.0/67.6 & 73.7/70.5 & 74.1/69.8 & 71.6/63.3 & 78.1/67.6 \\
Qwen3.5 Omni & 66.9/66.2 & 68.0/62.9 & 60.4/60.1 & 67.6/67.3 & 64.7/62.9 & 66.2/66.9 & 67.3/64.4 & 64.4/66.2 & 63.7/58.6 & 66.9/64.0 \\
Muse-Spark 1.2 & 86.3/66.9 & 75.9/43.2 & 81.2/36.1 & 83.8/56.5 & 83.8/55.4 & 83.1/56.1 & 79.1/55.4 & 83.8/54.7 & 77.0/37.8 & 83.0/43.5 \\
\addlinespace[3pt]
\midrule
\multicolumn{11}{@{}l}{\textbf{Self Coherence (SC)}} \\
\textbf{Model} & \textbf{EN} & \textbf{ZH} & \textbf{JA} & \textbf{ES} & \textbf{FR} & \textbf{RU} & \textbf{DE} & \textbf{IT} & \textbf{TH} & \textbf{VI} \\
\midrule
Qwen3.8 Omni & 80.1/82.3 & 80.4/81.4 & 75.7/77.9 & 83.9/83.3 & 81.1/80.4 & 80.4/80.4 & 84.9/81.1 & 76.7/80.4 & 80.1/77.9 & 81.1/83.0 \\
Gemini 3.1 Pro & 78.5/76.0 & 79.5/79.8 & 80.8/81.1 & 80.8/77.6 & 76.0/74.8 & 79.2/81.7 & 80.1/77.0 & 76.7/77.9 & 75.4/70.7 & 76.0/76.0 \\
Gemini 3.8 Flash & 76.7/72.9 & 75.7/77.0 & 76.0/71.0 & 74.4/76.7 & 75.4/75.1 & 78.2/77.3 & 76.7/76.3 & 73.5/70.3 & 75.1/72.2 & 74.1/72.9 \\
Doubao Seed Lite & 70.0/72.2 & 68.5/73.5 & 74.8/71.6 & 72.2/64.0 & 67.8/67.2 & 72.9/69.4 & 65.3/66.2 & 70.0/67.8 & 69.1/64.0 & 74.4/66.6 \\
Qwen3.5 Omni & 77.0/77.0 & 71.9/78.2 & 69.1/71.0 & 70.7/72.6 & 73.2/73.5 & 71.9/73.5 & 71.6/74.8 & 69.1/69.7 & 68.5/70.3 & 77.0/69.7 \\
Muse-Spark 1.2 & 82.0/76.3 & 84.9/73.5 & 82.6/67.2 & 80.8/70.3 & 80.4/74.1 & 85.5/81.7 & 85.2/78.2 & 82.3/75.4 & 77.0/69.4 & 81.1/69.1 \\
\addlinespace[3pt]
\midrule
\multicolumn{11}{@{}l}{\textbf{Voice Editing (VE)}} \\
\textbf{Model} & \textbf{EN} & \textbf{ZH} & \textbf{JA} & \textbf{ES} & \textbf{FR} & \textbf{RU} & \textbf{DE} & \textbf{IT} & \textbf{TH} & \textbf{VI} \\
\midrule
Qwen3.8 Omni & 89.9/90.9 & 91.3/89.6 & 90.5/90.4 & 92.1/90.4 & 91.6/88.4 & 90.5/92.1 & 90.7/89.5 & 89.3/88.0 & 88.1/88.3 & 91.9/87.2 \\
Gemini 3.1 Pro & 91.6/91.3 & 92.7/87.0 & 93.0/90.2 & 92.5/89.9 & 92.4/90.1 & 91.0/90.1 & 92.5/89.9 & 92.2/89.3 & 89.8/87.8 & 91.8/88.9 \\
Gemini 3.8 Flash & 90.7/87.7 & 92.8/89.2 & 90.7/88.6 & 91.3/87.8 & 88.6/86.1 & 89.9/88.4 & 89.2/87.2 & 90.4/87.5 & 89.5/85.5 & 91.0/84.6 \\
Doubao Seed Lite & 85.1/87.7 & 86.0/83.5 & 84.9/82.0 & 87.3/83.2 & 85.8/84.1 & 84.9/81.4 & 85.5/82.5 & 85.5/84.6 & 85.4/81.4 & 86.9/81.9 \\
Qwen3.5 Omni & 83.5/84.1 & 85.4/82.5 & 80.6/79.6 & 83.5/80.8 & 82.0/81.2 & 81.6/81.1 & 81.1/79.6 & 82.3/82.6 & 79.1/80.5 & 83.4/79.6 \\
Muse-Spark 1.2 & 91.0/67.1 & 91.2/56.2 & 90.5/61.7 & 91.5/69.8 & 92.2/67.4 & 91.8/67.2 & 91.6/65.0 & 90.1/71.8 & 91.8/60.2 & 89.8/60.1 \\
\bottomrule
\end{tabular*}
\endgroup
\end{table}

\subsection{Paired Capability Outcomes}

The paired rubric counts and survival rates behind Figure~\ref{fig:paired_capabilities} appear in Table~\ref{tab:paired_counts}, pooled over all ten languages.

% Generated by paper/tools/update_experiment_results.py from results/exp.md.
\begin{table}[htbp]
\centering
\caption{Paired rubric outcomes pooled over all ten languages (English human recordings plus nine non-English). Both, R only, A only, and neither refer to Reference Transcript / Direct Audio success. Survival conditions on reference-transcript success, so its denominator differs across models.}
\label{tab:paired_counts}
\begingroup
\fontsize{8.5}{10.2}\selectfont
\setlength{\tabcolsep}{1pt}
\renewcommand{\arraystretch}{1.06}
\begin{tabular*}{\linewidth}{@{}l@{\hspace{6pt}}l@{\extracolsep{\fill}}rrrrrr@{}}
\toprule
\textbf{Model} & \textbf{Capability} & \textbf{Both} & \textbf{R only} & \textbf{A only} & \textbf{Neither} & \textbf{N} & \textbf{Surv.\%} \\
\midrule
Qwen3.8 Omni & IM & 1023 & 134 & 95 & 278 & 1530 & 88.4 \\
 & IR & 2213 & 168 & 182 & 213 & 2776 & 92.9 \\
 & SC & 2370 & 180 & 192 & 428 & 3170 & 92.9 \\
 & VE & 5664 & 280 & 205 & 411 & 6560 & 95.3 \\
 & Overall & 11270 & 762 & 674 & 1330 & 14036 & 93.7 \\
\addlinespace[3pt]
Gemini 3.1 Pro & IM & 942 & 238 & 72 & 278 & 1530 & 79.8 \\
 & IR & 1980 & 415 & 106 & 279 & 2780 & 82.7 \\
 & SC & 2298 & 184 & 151 & 537 & 3170 & 92.6 \\
 & VE & 5700 & 332 & 169 & 359 & 6560 & 94.5 \\
 & Overall & 10920 & 1169 & 498 & 1453 & 14040 & 90.3 \\
\addlinespace[3pt]
Gemini 3.8 Flash & IM & 926 & 219 & 67 & 318 & 1530 & 80.9 \\
 & IR & 1830 & 283 & 155 & 512 & 2780 & 86.6 \\
 & SC & 2150 & 246 & 201 & 573 & 3170 & 89.7 \\
 & VE & 5510 & 421 & 214 & 415 & 6560 & 92.9 \\
 & Overall & 10416 & 1169 & 637 & 1818 & 14040 & 89.9 \\
\addlinespace[3pt]
Doubao Seed Lite & IM & 850 & 221 & 104 & 355 & 1530 & 79.4 \\
 & IR & 1760 & 329 & 164 & 527 & 2780 & 84.3 \\
 & SC & 1947 & 288 & 217 & 718 & 3170 & 87.1 \\
 & VE & 5167 & 457 & 293 & 643 & 6560 & 91.9 \\
 & Overall & 9724 & 1295 & 778 & 2243 & 14040 & 88.2 \\
\addlinespace[3pt]
Qwen3.5 Omni & IM & 693 & 153 & 144 & 540 & 1530 & 81.9 \\
 & IR & 1555 & 269 & 223 & 733 & 2780 & 85.3 \\
 & SC & 2085 & 197 & 230 & 658 & 3170 & 91.4 \\
 & VE & 4969 & 427 & 355 & 809 & 6560 & 92.1 \\
 & Overall & 9302 & 1046 & 952 & 2740 & 14040 & 89.9 \\
\addlinespace[3pt]
Muse-Spark 1.2 & IM & 747 & 472 & 67 & 238 & 1524 & 61.3 \\
 & IR & 1290 & 977 & 112 & 396 & 2775 & 56.9 \\
 & SC & 2203 & 402 & 128 & 437 & 3170 & 84.6 \\
 & VE & 4035 & 1896 & 175 & 404 & 6510 & 68.0 \\
 & Overall & 8275 & 3747 & 482 & 1475 & 13979 & 68.8 \\
\addlinespace[3pt]
\bottomrule
\end{tabular*}
\endgroup
\end{table}

\subsection{Additional Transcript-Source Diagnostics}

\label{app:transcript_sources}

The core $2\times2$ analysis varies the placement of transcript substitutions while holding their source fixed. This section instead examines how transcript source and reference-history controls change the observed results. These diagnostics are auxiliary to the main conditions: they alter the transcript-generation mechanism or replace speech history with the benchmark reference, and therefore do not belong to the factorial intervention.

\paragraph{Shared-recognizer and model-generated transcripts.}
Full Transcript, History Transcript, and Current Transcript use transcripts generated by the evaluated model. As a source diagnostic, we also replace all user turns with transcripts from a shared Gemini 3.5 Flash recognizer. Muse improves from \resMuseAudio\% under Direct Audio to \resMuseShared\% with the shared recognizer and \resMuseSelf\% with its model-generated Full Transcript input. Source effects are not uniform across models and languages, so this comparison does not identify one source as generally superior. Gemini 3.8 Flash uses fallback text for 3.1\% of model-generated transcription cases; its Full Transcript condition is consequently not a pure self-generated-transcript condition.

\paragraph{Recovery among identifiable Direct-Audio failures.}
Among instances that succeed under Reference Transcript and fail under Direct Audio, Full Transcript recovers \resMuseSelfRecovered/\resMuseSelfLost\ (\resMuseSelfRecovery\%) for Muse and \resGProSelfRecovered/\resGProSelfLost\ (\resGProSelfRecovery\%) for Gemini 3.1 Pro. These counts connect transcript-mediated improvements to previously lost reference successes, but they are subset-specific and do not account for newly introduced failures. Qwen3.8, for example, can recover some such failures even though its average Full Transcript contrast is negative. Table~\ref{tab:recovery_counts} reports the corresponding language-level counts and makes clear that eligible sets may differ across procedures.

\paragraph{Reference-history control.}
Replacing preceding audio with the Reference Transcript while retaining the current turn as audio provides a reference-history control. Its score is higher than the corresponding History Transcript score for all six models, consistent with sensitivity to transcript quality or source. This control is not an empirical upper bound on the Full Transcript condition: it changes only history, and for Muse Full Transcript is higher. Together, these diagnostics separate source and representation questions from the core localization result while leaving the mechanism of the observed improvements unresolved.

\begin{figure}[htbp]

\centering

\includegraphics[width=\linewidth]{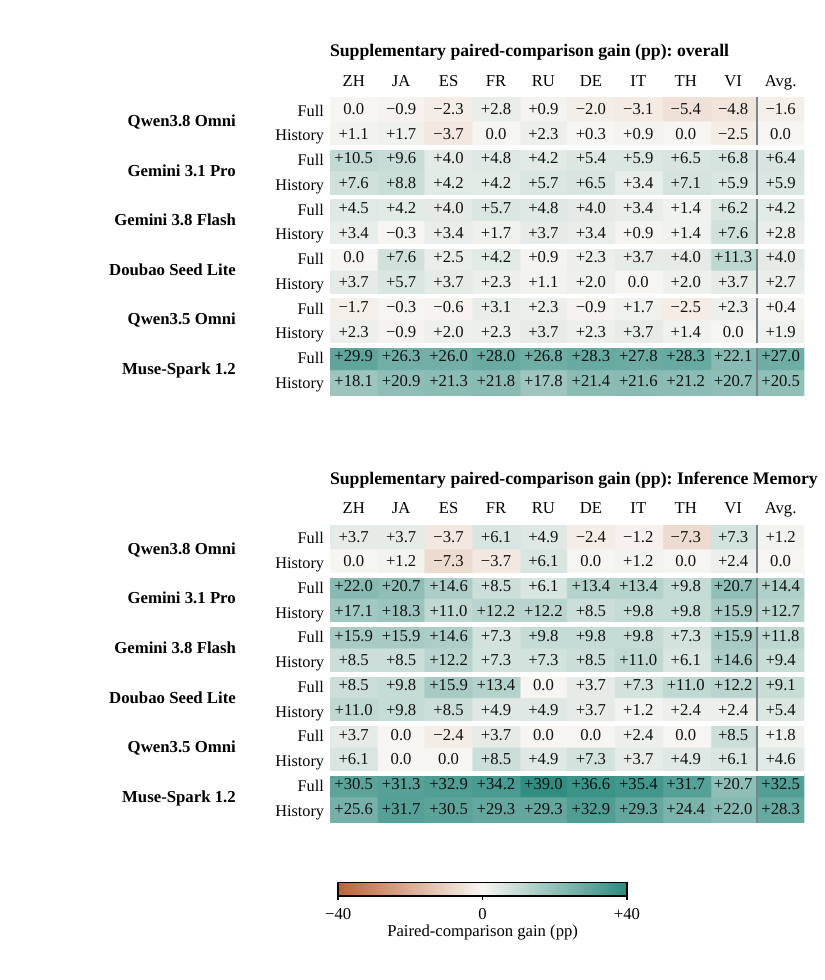}

\caption{Full-model comparison of Full Transcript and History Transcript paired instance-level gains over Direct Audio. Each model has Full and History rows; the two panels show overall and IM gains. Each procedure uses its own common samples and baseline. Avg. is the equal-weight mean of the nine non-English language gains (the paired export has no English arm). Signed values and an orange--teal scale distinguish losses and gains. These pairwise estimates are separate from the four-condition marginal contrasts in Table~\ref{tab:input_conditions}. These are descriptive point estimates.}

\label{fig:all_transcription_gains}

\end{figure}

For these diagnostics, Table~\ref{tab:source_conditions} reports the condition-level scores, Figure~\ref{fig:all_transcription_gains} compares Full and History paired gains for all six models, and Table~\ref{tab:recovery_counts} gives instance-level recovery counts.

% Generated by paper/tools/update_experiment_results.py from results/exp.md.
\begin{table}[htbp]
\centering
\caption{Additional transcript-source diagnostics: rubric / instance success (\%) by language. Full Transcript uses model-generated transcripts; Shared uses a common recognizer; Full denotes Full Transcript; Ref.hist. uses reference history and retains current audio. Avg. is the ten-language macro mean. All entries are marginal scores.}
\label{tab:source_conditions}
\begingroup
\fontsize{7.0}{9.0}\selectfont
\setlength{\tabcolsep}{1pt}
\renewcommand{\arraystretch}{1.06}
\begin{tabular*}{\linewidth}{@{}ll@{\extracolsep{\fill}}*{11}{r}@{}}
\toprule
\textbf{Model} & \textbf{Source} & \textbf{Avg.} & \textbf{EN} & \textbf{ZH} & \textbf{JA} & \textbf{ES} & \textbf{FR} & \textbf{RU} & \textbf{DE} & \textbf{IT} & \textbf{TH} & \textbf{VI} \\
\midrule
Qwen3.8 Omni & Shared & 84.4/59.6 & 84.5/59.6 & 83.7/57.0 & 84.5/59.9 & 86.7/63.3 & 85.9/62.7 & 85.3/64.1 & 86.5/62.7 & 84.4/59.6 & 80.6/51.1 & 82.3/55.6 \\
 & Full & 84.5/58.8 & 85.9/62.4 & 85.2/60.7 & 85.1/59.4 & 86.4/61.9 & 86.5/62.4 & 85.6/61.9 & 85.2/59.6 & 83.0/57.6 & 78.6/48.6 & 83.6/53.4 \\
 & Ref.hist. & 85.8/61.4 & 86.3/61.6 & 85.4/62.7 & 85.6/61.3 & 87.0/62.1 & 85.6/62.4 & 86.5/63.6 & 87.0/64.4 & 86.8/64.4 & 82.4/52.8 & 84.9/58.8 \\
\addlinespace[2pt]
Gemini 3.1 Pro & Shared & 84.7/59.4 & 85.3/61.6 & 85.5/60.5 & 83.5/57.3 & 86.1/61.9 & 84.8/60.7 & 86.0/62.4 & 84.9/57.9 & 85.0/59.6 & 82.1/57.1 & 83.4/54.8 \\
 & Full & 84.6/60.8 & 88.2/66.9 & 84.5/61.3 & 85.3/62.7 & 85.4/61.6 & 84.3/59.6 & 85.0/60.7 & 85.0/59.0 & 83.6/61.3 & 81.3/56.8 & 83.8/58.2 \\
 & Ref.hist. & 85.1/61.1 & 86.6/62.7 & 86.0/59.9 & 85.6/63.3 & 85.6/61.6 & 84.2/59.9 & 85.2/61.9 & 84.6/61.9 & 84.5/59.6 & 83.5/60.5 & 85.3/59.3 \\
\addlinespace[2pt]
Gemini 3.8 Flash & Shared & 81.5/54.0 & 82.2/55.1 & 82.4/56.1 & 81.3/53.7 & 82.6/55.6 & 81.9/54.7 & 82.1/55.6 & 80.6/52.5 & 82.1/55.4 & 78.3/49.2 & 81.3/52.3 \\
 & Full & 81.4/54.4 & 83.3/57.8 & 82.4/56.2 & 80.8/52.8 & 83.0/55.4 & 81.4/54.5 & 83.4/57.9 & 81.6/54.5 & 80.1/54.8 & 77.6/49.4 & 79.9/50.6 \\
 & Ref.hist. & 80.8/54.3 & 82.7/56.5 & 80.6/54.5 & 79.8/54.0 & 83.0/56.8 & 80.2/52.0 & 81.8/53.4 & 80.6/55.1 & 80.6/55.6 & 78.3/53.1 & 80.8/52.3 \\
\addlinespace[2pt]
Doubao Seed Lite & Shared & 77.4/46.2 & 78.1/48.0 & 76.7/46.3 & 76.8/47.5 & 78.7/47.2 & 77.1/44.9 & 80.3/50.6 & 78.6/46.3 & 76.6/45.5 & 74.6/41.5 & 76.9/44.4 \\
 & Full & 77.4/46.3 & 79.9/50.3 & 78.6/46.3 & 77.1/49.7 & 75.6/42.9 & 77.4/45.5 & 76.8/46.0 & 76.7/43.8 & 78.3/48.3 & 75.8/42.7 & 77.3/47.2 \\
 & Ref.hist. & 78.3/48.0 & 80.0/52.0 & 79.0/48.9 & 79.3/52.5 & 79.3/50.0 & 78.6/46.9 & 77.3/48.0 & 78.8/48.0 & 78.3/46.9 & 75.9/43.8 & 76.5/42.9 \\
\addlinespace[2pt]
Qwen3.5 Omni & Shared & 72.9/38.8 & 74.0/38.4 & 73.4/39.8 & 70.1/40.1 & 73.7/37.0 & 74.1/42.1 & 74.9/41.0 & 74.0/40.1 & 73.1/39.3 & 68.5/32.5 & 73.2/38.1 \\
 & Full & 72.9/38.1 & 75.5/39.5 & 73.2/38.1 & 71.1/38.4 & 72.5/37.9 & 73.2/40.4 & 74.6/40.7 & 73.8/36.7 & 73.3/40.1 & 69.5/33.6 & 72.6/35.9 \\
 & Ref.hist. & 73.8/40.5 & 76.4/43.2 & 74.6/43.5 & 70.6/40.1 & 74.1/40.7 & 74.8/41.2 & 75.1/44.9 & 74.3/40.7 & 73.8/38.4 & 70.0/34.5 & 74.1/37.6 \\
\addlinespace[2pt]
Muse-Spark 1.2 & Shared & 85.2/61.9 & 85.8/64.4 & 85.1/61.4 & 85.4/62.0 & 85.8/59.6 & 84.3/61.6 & 88.0/66.7 & 85.3/62.7 & 85.5/62.1 & 82.2/58.2 & 84.9/60.3 \\
 & Full & 83.2/58.6 & 83.0/58.5 & 83.5/58.6 & 82.2/55.6 & 84.6/60.2 & 82.5/60.5 & 85.4/64.7 & 85.0/61.9 & 84.5/60.2 & 79.8/55.4 & 81.2/50.3 \\
 & Ref.hist. & 81.0/54.8 & 82.2/57.1 & 78.3/52.0 & 79.7/56.9 & 83.6/57.3 & 83.5/57.3 & 82.0/55.2 & 83.5/57.9 & 81.4/55.1 & 75.9/47.7 & 79.8/51.1 \\
\addlinespace[2pt]
\bottomrule
\end{tabular*}
\endgroup
\end{table}

% Generated by paper/tools/update_experiment_results.py from results/exp.md.
\begin{table}[htbp]
\centering
\caption{Instance-level recovery among reference-transcript successes that fail under direct audio. Each cell gives recovered / eligible cases; the final column pools available language counts. Full/Hist./Ref.hist. denote Full Transcript, History Transcript, and reference-history input. Eligible sets can differ between procedures. Recovery does not account for new failures.}
\label{tab:recovery_counts}
\begingroup
\fontsize{7.8}{9.5}\selectfont
\setlength{\tabcolsep}{1pt}
\renewcommand{\arraystretch}{1.06}
\begin{tabular*}{\linewidth}{@{}l@{\hspace{5pt}}l@{\extracolsep{\fill}}*{9}{r}r@{}}
\toprule
\textbf{Model} & \textbf{Input} & \textbf{ZH} & \textbf{JA} & \textbf{ES} & \textbf{FR} & \textbf{RU} & \textbf{DE} & \textbf{IT} & \textbf{TH} & \textbf{VI} & \textbf{Pooled\%} \\
\midrule
Muse-Spark 1.2 & Full & 104/140 & 93/133 & 92/126 & 103/131 & 98/125 & 99/125 & 103/126 & 104/136 & 87/129 & 75.4 \\
 & Hist. & 77/140 & 70/131 & 86/126 & 88/131 & 77/125 & 89/124 & 88/126 & 88/136 & 79/129 & 63.5 \\
 & Ref.hist. & 92/141 & 93/130 & 91/126 & 95/131 & 85/125 & 86/125 & 96/126 & 77/136 & 90/129 & 68.9 \\
Gemini 3.1 Pro & Full & 50/71 & 47/64 & 29/49 & 36/50 & 28/45 & 38/52 & 39/51 & 44/56 & 36/55 & 70.4 \\
 & Hist. & 43/71 & 45/64 & 28/49 & 35/50 & 28/45 & 29/52 & 33/51 & 40/56 & 39/55 & 64.9 \\
 & Ref.hist. & 49/71 & 43/64 & 32/49 & 37/50 & 29/45 & 28/52 & 36/51 & 40/56 & 37/55 & 67.1 \\
Qwen3.8 Omni & Full & 18/34 & 17/30 & 20/42 & 19/38 & 26/44 & 23/43 & 18/36 & 10/32 & 24/43 & 51.2 \\
 & Hist. & 19/34 & 17/30 & 16/42 & 20/38 & 26/44 & 24/43 & 18/36 & 17/32 & 17/43 & 50.9 \\
 & Ref.hist. & 22/34 & 15/30 & 20/42 & 28/38 & 28/44 & 22/43 & 24/36 & 14/32 & 17/43 & 55.6 \\
Gemini 3.8 Flash & Full & 30/48 & 38/57 & 25/44 & 30/48 & 24/44 & 36/54 & 31/48 & 26/47 & 39/66 & 61.2 \\
 & Hist. & 26/48 & 28/57 & 22/44 & 27/48 & 24/44 & 31/54 & 25/48 & 25/47 & 33/66 & 52.9 \\
 & Ref.hist. & 28/48 & 36/57 & 20/44 & 33/48 & 23/44 & 36/54 & 30/48 & 25/47 & 33/66 & 57.9 \\
Doubao Seed Lite & Full & 22/44 & 35/50 & 21/54 & 28/44 & 26/48 & 24/48 & 23/42 & 27/47 & 51/80 & 56.2 \\
 & Hist. & 21/44 & 35/50 & 21/54 & 18/44 & 23/48 & 27/48 & 17/42 & 25/47 & 36/80 & 48.8 \\
 & Ref.hist. & 26/44 & 35/50 & 32/54 & 24/44 & 27/48 & 32/48 & 26/42 & 25/47 & 50/80 & 60.6 \\
Qwen3.5 Omni & Full & 14/33 & 18/36 & 14/35 & 23/33 & 18/36 & 15/36 & 18/31 & 13/30 & 22/42 & 49.7 \\
 & Hist. & 11/33 & 17/36 & 18/35 & 18/33 & 19/36 & 15/36 & 14/31 & 14/30 & 17/42 & 45.8 \\
 & Ref.hist. & 20/33 & 18/36 & 19/35 & 19/33 & 23/36 & 22/36 & 16/31 & 12/30 & 19/42 & 53.8 \\
\bottomrule
\end{tabular*}
\endgroup
\end{table}

\subsection{English Speech and History Controls}

\label{app:history_controls}

% Generated by paper/tools/update_experiment_results.py from results/exp.md.
\begin{table}[htbp]
\centering
\caption{English human/TTS control. Entries are rubric / instance success (\%). Both speech versions share the same English reference transcript. The final column uses reference-transcript history with a current TTS turn. Synthetic English is a control condition, not an additional language.}
\label{tab:english_control}
\begingroup
\fontsize{8.5}{10.2}\selectfont
\setlength{\tabcolsep}{1pt}
\renewcommand{\arraystretch}{1.06}
\begin{tabular*}{\linewidth}{@{}l@{\extracolsep{\fill}}rrrrr@{}}
\toprule
\textbf{Model} & \textbf{Human audio} & \textbf{TTS audio} & \textbf{TTS shared} & \textbf{TTS self} & \textbf{TTS clean hist.} \\
\midrule
Qwen3.8 Omni & 86.8/61.6 & 85.5/60.2 & 84.9/60.5 & 84.9/60.2 & 86.0/60.6 \\
Gemini 3.1 Pro & 84.1/60.2 & 85.0/60.7 & 84.1/59.6 & 86.0/60.7 & 86.0/61.3 \\
Gemini 3.8 Flash & 80.3/52.0 & 79.7/50.6 & 81.4/53.4 & 83.0/55.7 & 81.4/54.2 \\
Doubao Seed Lite & 79.3/47.7 & 77.4/47.5 & 77.6/46.3 & 78.4/45.2 & 79.5/48.6 \\
Qwen3.5 Omni & 75.9/42.9 & 73.4/38.4 & 72.4/39.8 & 75.6/42.9 & 76.1/45.2 \\
Muse-Spark 1.2 & 68.7/39.8 & 73.0/39.6 & 86.0/62.7 & 84.7/61.0 & 84.6/58.9 \\
\bottomrule
\end{tabular*}
\endgroup
\end{table}

% Generated by paper/tools/update_experiment_results.py from results/exp.md.

\section{Locally Evaluated Open Models}

\label{app:open_models}

% Generated by paper/tools/update_experiment_results.py from results/exp.md.
\begin{table}[htbp]
\centering
\caption{Local open models: ten-language macro-mean \textbf{rubric / instance} success (\%), English human recordings included. Scores use condition-specific retained sets; coverage is in Table~\ref{tab:open_coverage}.}
\label{tab:open_summary}
\begingroup
\fontsize{8.5}{10.5}\selectfont
\setlength{\tabcolsep}{1pt}
\renewcommand{\arraystretch}{1.06}
\begin{tabular*}{\linewidth}{@{}l@{\extracolsep{\fill}}rr@{}}
\toprule
\textbf{Model} & \textbf{Reference Transcript} & \textbf{Direct Audio} \\
\midrule
Qwen2.5-Omni & 47.5/13.8 & 44.1/12.9 \\
Voxtral-Mini & 57.6/20.1 & 56.1/18.5 \\
Qwen3-Omni & 66.5/29.0 & 62.4/24.4 \\
Voxtral-Small & 69.2/30.4 & 66.8/28.1 \\
\bottomrule
\end{tabular*}
\endgroup
\end{table}

Four additional models are evaluated locally on the Reference and Audio conditions using the same o4-mini judge and evaluation rubrics. Conversations that embed background sounds and incomplete conversations are excluded before scoring. Table~\ref{tab:open_summary} summarizes their ten-language macro scores, Table~\ref{tab:open_languages} reports language-level scores, Table~\ref{tab:open_capabilities} capability-level rates, and Table~\ref{tab:open_english} the English controls. Retained sets differ across languages and conditions, so these results extend benchmark coverage without supplying paired modality-transition evidence. No shared-ASR, self-ASR, or hybrid results are available for these models. Reported model names follow the evaluation export.

% TODO(open-model reproducibility): Specify exact checkpoints, parameter counts, inference framework, precision, decoding settings, and completion/filtering criteria before submission.

% Generated by paper/tools/update_experiment_results.py from results/exp.md.
\begin{table}[htbp]
\centering
\caption{Language-level Reference and Audio scores for the locally evaluated open models. Each cell is rubric / instance success (\%); Avg. is the ten-language macro mean (English human recordings included). Conversation counts appear in Table~\ref{tab:open_coverage}.}
\label{tab:open_languages}
\begingroup
\fontsize{7.2}{9.2}\selectfont
\setlength{\tabcolsep}{1pt}
\renewcommand{\arraystretch}{1.06}
\begin{tabular*}{\linewidth}{@{}l@{\hspace{5pt}\extracolsep{\fill}}*{11}{r}@{}}
\toprule
\multicolumn{12}{@{}l}{\textbf{Direct Audio}} \\
\textbf{Model} & \textbf{Avg.} & \textbf{EN} & \textbf{ZH} & \textbf{JA} & \textbf{ES} & \textbf{FR} & \textbf{RU} & \textbf{DE} & \textbf{IT} & \textbf{TH} & \textbf{VI} \\
\midrule
Qwen2.5-Omni & 44.1/12.9 & 51.7/14.5 & 49.9/15.0 & 40.4/12.6 & 45.8/12.9 & 45.8/13.1 & 44.2/12.6 & 46.9/14.1 & 42.6/10.3 & 34.0/11.4 & 40.0/12.5 \\
Voxtral-Mini & 56.1/18.5 & 61.0/22.4 & 54.1/16.3 & 52.6/16.2 & 59.7/22.1 & 59.2/20.4 & 57.3/18.8 & 58.6/18.7 & 57.2/16.7 & 47.4/14.4 & 54.0/19.0 \\
Qwen3-Omni & 62.4/24.4 & 65.1/28.2 & 65.2/28.0 & 60.8/23.2 & 64.4/24.4 & 63.7/26.0 & 61.5/22.0 & 65.9/30.2 & 60.0/22.9 & 57.6/18.4 & 60.0/20.6 \\
Voxtral-Small & 66.8/28.1 & 68.6/27.4 & 64.9/26.3 & 64.6/27.2 & 69.9/29.4 & 68.3/29.4 & 68.0/31.6 & 68.9/27.8 & 69.5/31.4 & 62.1/24.9 & 62.8/25.2 \\
\addlinespace[5pt]
\midrule
\multicolumn{12}{@{}l}{\textbf{Reference Transcript}} \\
\textbf{Model} & \textbf{Avg.} & \textbf{EN} & \textbf{ZH} & \textbf{JA} & \textbf{ES} & \textbf{FR} & \textbf{RU} & \textbf{DE} & \textbf{IT} & \textbf{TH} & \textbf{VI} \\
\midrule
Qwen2.5-Omni & 47.5/13.8 & 52.5/16.4 & 48.6/12.5 & 43.3/14.2 & 49.0/13.4 & 49.6/16.1 & 47.9/12.3 & 48.6/13.5 & 47.0/11.8 & 41.4/12.6 & 47.1/14.8 \\
Voxtral-Mini & 57.6/20.1 & 62.3/25.1 & 59.4/19.5 & 57.6/20.5 & 59.0/20.9 & 61.6/23.3 & 55.3/19.5 & 59.2/21.6 & 56.2/15.9 & 50.1/15.0 & 55.5/19.6 \\
Qwen3-Omni & 66.5/29.0 & 66.4/28.2 & 70.2/36.2 & 66.4/28.9 & 67.8/29.9 & 67.2/28.6 & 65.9/27.7 & 65.7/29.4 & 65.0/25.4 & 62.4/25.2 & 68.2/30.2 \\
Voxtral-Small & 69.2/30.4 & 70.3/29.0 & 70.9/29.9 & 68.4/29.7 & 72.2/33.6 & 70.2/34.5 & 69.0/31.6 & 67.8/26.6 & 72.2/35.6 & 62.5/23.7 & 68.7/29.7 \\
\bottomrule
\end{tabular*}
\endgroup
\end{table}

% Generated by paper/tools/update_experiment_results.py from results/exp.md.
\begin{table}[htbp]
\centering
\caption{Capability-level rubric success (\%), averaged equally across the nine non-English languages (no per-language English capability export). Each cell reports Reference Transcript / Direct Audio; these are marginal rates, not survival probabilities.}
\label{tab:open_capabilities}
\begingroup
\fontsize{8.5}{10.2}\selectfont
\setlength{\tabcolsep}{1pt}
\renewcommand{\arraystretch}{1.06}
\begin{tabular*}{\linewidth}{@{}l@{\extracolsep{\fill}}rrrr@{}}
\toprule
\textbf{Model} & \textbf{IM} & \textbf{IR} & \textbf{SC} & \textbf{VE} \\
\midrule
Qwen2.5-Omni & 31.6/30.5 & 30.4/31.3 & 51.3/47.8 & 55.4/49.2 \\
Voxtral-Mini & 34.6/32.8 & 42.6/41.6 & 63.2/61.8 & 65.5/63.7 \\
Qwen3-Omni & 43.1/39.4 & 54.7/48.8 & 67.3/64.5 & 76.6/72.0 \\
Voxtral-Small & 40.3/40.3 & 55.8/52.9 & 73.1/73.4 & 79.5/75.2 \\
\bottomrule
\end{tabular*}
\endgroup
\end{table}

\clearpage

% Generated by paper/tools/update_experiment_results.py from results/exp.md.
\begin{table}[htbp]
\centering
\caption{Scored conversations for the local models after exclusion of incomplete conversations. Each condition targets at most 354 conversations; differing counts do not identify the overlap of retained sets.}
\label{tab:open_coverage}
\begingroup
\fontsize{8}{9.6}\selectfont
\setlength{\tabcolsep}{1pt}
\renewcommand{\arraystretch}{1.06}
\begin{tabular*}{\linewidth}{@{}l@{\hspace{5pt}}l@{\extracolsep{\fill}}*{11}{r}@{}}
\toprule
\textbf{Model} & \textbf{Input} & \textbf{EN} & \textbf{EN-TTS} & \textbf{ZH} & \textbf{JA} & \textbf{ES} & \textbf{FR} & \textbf{RU} & \textbf{DE} & \textbf{IT} & \textbf{TH} & \textbf{VI} \\
\midrule
Qwen2.5-Omni & Reference Transcript & 353 & 353 & 352 & 353 & 352 & 347 & 351 & 348 & 347 & 340 & 352 \\
 & Direct Audio & 352 & 353 & 354 & 348 & 350 & 343 & 348 & 347 & 348 & 334 & 345 \\
\addlinespace[3pt]
Voxtral-Mini & Reference Transcript & 354 & 354 & 353 & 352 & 354 & 352 & 354 & 352 & 353 & 346 & 352 \\
 & Direct Audio & 353 & 354 & 350 & 352 & 353 & 353 & 352 & 353 & 354 & 334 & 347 \\
\addlinespace[3pt]
Qwen3-Omni & Reference Transcript & 354 & 354 & 354 & 353 & 354 & 353 & 354 & 354 & 354 & 353 & 354 \\
 & Direct Audio & 354 & 352 & 354 & 354 & 353 & 354 & 354 & 354 & 354 & 354 & 354 \\
\addlinespace[3pt]
Voxtral-Small & Reference Transcript & 352 & 352 & 354 & 354 & 354 & 354 & 354 & 354 & 354 & 350 & 353 \\
 & Direct Audio & 354 & 354 & 353 & 353 & 354 & 354 & 354 & 353 & 353 & 350 & 353 \\
\addlinespace[3pt]
\bottomrule
\end{tabular*}
\endgroup
\end{table}

% Generated by paper/tools/update_experiment_results.py from results/exp.md.
\begin{table}[htbp]
\centering
\caption{English controls for the local models, shown as rubric / instance success (\%). Unlike the primary evaluation, the EN-TTS reference condition is an independent run; differences between its reference score and EN reference must not be attributed to speech synthesis.}
\label{tab:open_english}
\begingroup
\fontsize{8.5}{10.2}\selectfont
\setlength{\tabcolsep}{1pt}
\renewcommand{\arraystretch}{1.06}
\begin{tabular*}{\linewidth}{@{}l@{\extracolsep{\fill}}rrrr@{}}
\toprule
\textbf{Model} & \textbf{EN audio} & \textbf{EN-TTS audio} & \textbf{EN ref.} & \textbf{EN-TTS ref.} \\
\midrule
Qwen2.5-Omni & 51.7/14.5 & 49.3/13.9 & 52.5/16.4 & 52.0/14.7 \\
Voxtral-Mini & 61.0/22.4 & 61.0/22.9 & 62.3/25.1 & 61.8/27.7 \\
Qwen3-Omni & 65.1/28.3 & 66.6/27.8 & 66.4/28.3 & 66.8/28.3 \\
Voxtral-Small & 68.6/27.4 & 68.3/27.7 & 70.3/29.0 & 70.6/31.8 \\
\bottomrule
\end{tabular*}
\endgroup
\end{table}

\section{Construction Details}
\label{app:construction_details}

\subsection{Localization and Text Validation}

Transcript analysis extracts semantic content, final intent, and disfluency annotations from the English conversations. Each user turn is localized using the English source history and previously localized target-language turns to maintain cross-turn consistency. Sequential user-turn localization produces a spoken-form target, an auxiliary fluent rendering, localized withdrawn and final values, and delivery guidance. The spoken-form target supplies both the TTS content and the reference-transcript evaluation input. The fluent rendering resolves repairs and removes disfluencies; it supports semantic comparison of the resolved content during construction, with disfluency structure verified against the spoken-form target by the structural checks, and it is not used for synthesis or as the reference-transcript evaluation condition. Structural checks verify annotated critical values in the spoken text and distinguish their ordered occurrences when a repair returns to an earlier value. The core localization prompt, which renders each user turn as speech in the target language while preserving disfluency and repair structure, is reproduced in Listing~\ref{lst:localization_prompt}.

\begin{lstlisting}[caption={The one-step spoken-localization prompt (Stage 2, Route A), used verbatim in construction.}, label={lst:localization_prompt}]
# Stage 2 / Route A -- One-Step Spoken Localization

You re-say one turn of a spoken conversation in a different language, **as speech, not as prose**. The speaker in the output must stumble, hesitate, and correct themselves in the same places and for the same reasons as the speaker in the input.

Target language: `{{TARGET_LANGUAGE}}`

## The one mistake that ruins this task

You are a translation model, so your instinct is to produce clean, well-formed output. Here that instinct is wrong. Consider this input:

> `the ground beef actually needs to be an 80%, weight, 85%, weight. 70.`

A fluent translation says *"70% ground beef"*. **That is a failure.** The benchmark item exists to test whether a listener can follow a live revision, so `80%` and `85%` must still be **audible** in your output, in the order they were abandoned, each marked as withdrawn the way a native speaker of `{{TARGET_LANGUAGE}}` would mark it. Delete a retracted value and the item becomes trivially easy -- the data is worthless.

## Non-negotiable rules

1. **Every value in `retracted` must appear in `spoken_target`.** Not paraphrased, not summarized -- spoken, then withdrawn.
2. **Every value in `final` must be the one that stands** at the end of the utterance.
3. **Preserve the order of revision.** If the speaker went `80% -> 85% -> 70%`, your output goes through the same three values in that sequence. Do not reorder, do not collapse.
4. **Keep the repair in the same clause.** A mid-sentence correction must stay mid-sentence. Do not tidy it into a separate follow-up sentence.
5. **Do not add information, opinions, or explanations** that the speaker did not utter.
6. **Do not translate word-for-word.** Say what a native speaker would actually say.

Two things to get right regardless of language:

- **Numerals and measure words.** When a number is respoken, it must carry whatever counter, classifier, case, or agreement `{{TARGET_LANGUAGE}}` requires in that slot -- including the retracted ones. A bare digit dropped into the sentence is wrong if the language would not permit it there.
- **Register.** The whole conversation is `{{REGISTER}}`. Use the politeness level, pronouns, and verb forms that a native speaker would sustain at that level, and keep it constant across every turn.

## Output format

Return **only** a JSON object, no prose, no markdown fence.

{
  "spoken_target": "The turn as actually spoken in {{TARGET_LANGUAGE}}: disfluent, self-correcting, natural. This is what will be sent to text-to-speech.",
  "clean_target": "The same content in fluent {{TARGET_LANGUAGE}}, all disfluency removed and all corrections already applied. Reference only; not synthesized.",
  "disfluency_target": [
    {
      "type": "SELF_CORRECTION",
      "surface_target": "verbatim substring of spoken_target that carries this repair",
      "retracted_target": ["verbatim substring of spoken_target for each abandoned value"],
      "final_target": "verbatim substring of spoken_target for the surviving value"
    }
  ],
  "style_instruction": "A short delivery instruction for the TTS engine, written in {{TARGET_LANGUAGE}}. Describe pace, hesitancy, and where the speaker breaks off. Do not restate the sentence itself.",
  "self_check": {
    "retracted_all_present": true,
    "final_values_correct": true,
    "notes": "Leave empty if clean. If you had to deviate from the source structure, say why in one sentence."
  }
}

### Before you answer

**Every string in `surface_target`, `retracted_target`, and `final_target` must be copied character-for-character out of `spoken_target`.** Do not summarize, join, or rephrase a value for the metadata -- a downstream step matches these strings literally against speech recognition output, so a paraphrase silently breaks it. If a repair spans two separate places in the utterance, pick the one contiguous substring that carries the value.

Then verify each of these against your own `spoken_target`:

- Every string in the input's `retracted` has a counterpart in `retracted_target`, and each of those counterparts literally occurs inside `spoken_target`.
- `final_target` occurs inside `spoken_target`, and it is the value a listener would act on.
- Reading `spoken_target` aloud sounds like a person talking, not a person reading.
- `clean_target` and `spoken_target` resolve to the same final intent.

If any check fails, fix `spoken_target` before returning. Set the corresponding `self_check` flag to `false` only if it is genuinely impossible in `{{TARGET_LANGUAGE}}`, and explain in `notes`.
\end{lstlisting}

The prompt additionally contains a table that illustrates, for a few languages, how each disfluency device (fillers, pauses, repetition, repair markers, and approximations) surfaces in native casual speech, together with an explicit instruction not to copy or transliterate the example tokens; the table's examples are typeset in target-language scripts and are omitted here. The template closes with three placeholders: the conversation context, the verbatim source turn, and a structural analysis of that turn produced by an earlier pipeline stage. The complete prompt files for all construction stages (turn analysis, spoken-form localization, delivery-style writing, rubric localization, transcription, and localization auditing) are released verbatim with the code.

Assistant turns and rubrics are localized in a conversation-level call, with checks for nonempty rubrics, preservation of rubric counts, and completeness of assistant turns. Text auditing presents source and target turns together with aligned rubrics, checking naturalness and language-dependent constraints such as word counts; rubric items that rely on language-specific form, such as an English word-count limit, are rewritten into a language-appropriate criterion or removed across parallel conditions when they cannot be transferred. Fillers and hesitation cues are assessed in context rather than by strict surface matching. User-turn issues are primarily reported rather than automatically rewritten; targeted terminology changes require review and resynthesis of affected audio. Rubric defects that may affect scoring are removed consistently across parallel conditions, with original rubric indices retained for alignment. Source-level inconsistencies are excluded rather than automatically corrected. The text audit also flags conversations with severe problems in at least two target languages for parallel removal.

\subsection{Synthesis Settings and Recovery}

Synthesis uses \texttt{gemini-3.1-flash-tts-preview} \citep{google2026gemini31flashtts}, an explicit language tag, and a frozen conversation-to-voice mapping drawn from 30 presets. The assignment is shared across synthesized languages and is not matched to the original English speaker's gender. Output audio is 24~kHz, mono, 16-bit PCM. Delivery guidance, derived from punctuation, discourse fillers, and conversational context, is supplied as a natural-language prefix. Recovery options include the original prefixed text, plain text, and a language-specific read-aloud frame. Read-aloud framing instructs the synthesizer to vocalize the supplied text rather than execute instruction-like content. Voice assignment remains fixed throughout recovery.

For \texttt{en\_tts}, the source English transcript is copied without translation or rewriting. Delivery instructions are generated separately in one call per conversation. This control therefore changes speech realization while retaining the English reference transcripts.

\subsection{Audio Content Review and Resynthesis}

Multiple ASR systems transcribe each utterance, and their outputs are compared with the disfluency-preserving spoken-form reference. The minimum normalized CER across recognizers prioritizes utterances for semantic review, reducing reliance on discrepancies found in a single transcription. It is used for screening rather than as a standalone acceptance criterion.

A language model receives the reference together with the available transcripts and assesses both missing reference content and unsupported additions or substitutions. Review distinguishes potential changes to names, quantities, negation, and other task-relevant content from spelling, transliteration, and numeral-format differences. Disfluency differences require contextual interpretation: variation in a filler may leave the task unchanged, whereas omission of a withdrawn value can remove a required revision even when the final intent is preserved. Transcription alone cannot determine whether such smoothing occurred in recognition or synthesis, so ambiguous cases are referred for further review.

Semantic review also extends beyond the screened subset to cover low-CER utterances. Targeted phonetic checks and manual review assess disputed cases, particularly where homophonic spellings produce different written forms for the same speech.

Utterances selected for repair are resynthesized from the same spoken-form reference with the assigned voice retained. Replacement candidates undergo renewed transcription and duration checks before acceptance. Review and repair are guided by content fidelity; their effect on task scores is assessed separately by re-evaluating affected conversations. Audio revisions preserve the shared reference transcripts and evaluation criteria across conditions. After iterative resynthesis and layered review, the final audio version carries no unresolved content defect identified by these checks.

The construction audit yields an explicit accounting. Parallel filtering removes 91 of the 452 source conversations---42 audio-cue conversations that embed background sounds, outside the speech-only scope of this work, and 49 removed for confirmed defects---together with their aligned versions in every language; seven further conversations are excluded by the realization-level parallel filtering of Section~\ref{sec:quality_control}, giving the final 354-conversation aligned set. Rubric items follow the same waterfall: of 1{,}712 localized items, 82 single-language translation defects, 20 source-defect items flagged independently in at least three languages, 5 language-inapplicable items, and 63 items of the audio-cue conversations are removed, and 122 further items are lost with their removed conversations (62 of them themselves defective); a further 16 items leave with the seven unrealizable conversations, giving the 1{,}404 items of the aligned set. Audio-level QC screens all 16{,}438 synthesized user turns with five-ASR evidence fusion, flagging a turn only when its minimum cross-ASR character error rate exceeds 5\%---well above the per-language median noise CER of about 2\%; of 521 flagged turns, semantic adjudication confirms 191 as true defects, and 176 are repaired and re-synthesized. Re-scoring the affected runs on the repaired audio changes instance success by at most 0.15 percentage points. Human validation draws 50 conversations per language (40 per-language samples plus 10 shared across all languages), supporting both an unconditional defect-rate estimate and direct cross-language comparison of the same content.

\subsection{Benchmark Validity and Robustness}
\label{sec:validity_results}

The construction checks above assess content fidelity; downstream robustness controls address whether synthesis and accepted audio repairs materially affect the evaluation.

The English human/TTS comparison provides a complementary control for synthesis sensitivity. To assess sensitivity to accepted audio repairs, affected conversation--language cases are re-evaluated under the same model and scoring settings, while unchanged cases retain their existing results. This comparison assesses the effect of audio revisions on benchmark scores; the transcript-mediated analysis separately tests whether an explicit transcript supports better use of the spoken content.

\section{Judge Robustness}
\label{app:judge_robustness}

To separate judge sensitivity from model behavior, we re-judge frozen responses with an alternative judge family (GPT-5.6-terra \citep{openai2026gpt56terra}) on byte-identical evaluation inputs, so score differences reflect the judge family rather than response variation. Estimates in this section are rubric-level pass rates on the stated subsets and are not directly comparable to the instance-level macro means in the main tables. The scoring prompt itself is reproduced verbatim at the end of this appendix.

Across the full Direct Audio condition for Gemini 3.1 Pro (15{,}444 paired rubric judgments over eleven language versions), the two judges agree on 94.5\% of judgments, the ten-language macro instance-success rates differ by 1.0 point, and the disagreements are nearly balanced between the alternative judge being stricter (419) and more lenient (429), indicating noise rather than directional bias.

On a capability-stratified panel (60 conversations, 15 per axis; three models; five conditions; up to eleven language versions; 37{,}826 paired rubric judgments), judgment agreement ranges from 93.4\% to 95.3\% across the fifteen model-by-condition cells, consistent with the full-set A/B above. The alternative judge is slightly more lenient on average (at most 2.6 rubric points per cell), and this leniency shows no systematic variation across conditions, so it largely cancels in contrasts: the two judges never differ by more than 1.2 points on any of the fifteen panel contrasts, well below the contrasts themselves. The headline contrasts are therefore stable under both judges: the Reference--Direct gap for Muse-Spark is 22.6 points under the primary judge versus 23.1 under the alternative, and its History--Current contrast is 13.1 versus 13.5. For the smaller contrasts (Gemini 3.1 Pro's History--Current contrast and all of Qwen3.8 Omni's contrasts, a few points or less on the full set), the two judges' panel estimates likewise agree within one point, so residual uncertainty about them reflects the 60-conversation panel's sampling noise rather than judge disagreement, and their full-set point estimates and intervals appear in the main tables and Appendix~\ref{app:bootstrap_ci}. Panel-level rubric rates are subsample estimates and are not comparable to the full-set instance-level values; instance-level panel contrasts are not reported because binary per-conversation outcomes make them unstable: the same Muse-Spark contrast estimated on two independently drawn panels (40 and 60 conversations) differs by more than 14 points, while rubric-level agreement remains stable. Same-judge repeated scoring is not run.

The scoring prompt is identical English for every language and every input condition, so the grading standard is comparable across languages; only the judged content (rubric items and conversation history) is localized. Long lines are wrapped for page width and non-ASCII punctuation is rendered with ASCII equivalents; the wording is unchanged from the prompt files used in the pipeline. The prompt follows the rubric-scoring template of the upstream AudioMC benchmark; the \texttt{conversation\_history} and \texttt{rubric\_item} placeholders are filled with the assembled conversation history and one rubric item, and the judge returns one JSON object per rubric item.

\begin{lstlisting}[caption={The rubric-scoring prompt, used verbatim for every judged condition.}, label={lst:judge_prompt}]
Your job is to look at a conversation and a rubric item, and score the last turn (i.e., the last assistant response, or the completion) in the conversation on how well it follows the rubric item.
# Conversation History
<<conversation_history>>
# Rubric item
<<rubric_item>>
# Instructions
Return a json object with the following fields: "explanation" and "criteria_met".
- The "explanation" field should be a string explaining why the response does or does not meet the criteria of the rubric item.
- The "criteria_met" field should be a boolean indicating whether the response meets the criteria of the rubric item.
# Example 1
For example, if the response is "For tonight's dinner, think of making a cheese and mushroom pizza" and the rubric item is "Avoids dairy product in the recipes", you should return a json like this:
```json
{
"explanation": "The response includes a dairy product in the recipe, so the criteria was not met.",
"criteria_met": false
}
```
# Final instruction
Return just the json object in markdown format. Do not include any other text in the response.
\end{lstlisting}

\section{Limitations}
\label{app:Limitations}
The fixed-context protocol does not evaluate online error accumulation, latency, or turn-taking. History/current interventions establish behavioral sensitivity to input representation without identifying a particular internal memory mechanism, and they replace different amounts of audio. The factorial contrasts in the main tables use marginal scores, and the common-set interaction intervals span or nearly span zero, so interaction size and stability remain unestablished. Transcription changes both context representation and computation, and we do not establish how errors scale with dialogue duration. Non-English speech is synthesized, and fixed voice assignments with English controls cannot establish equivalent synthesis effects across languages. Results remain conditional on scoring coverage and model-based judgments; an alternative judge family shows no systematic bias and reproduces the largest contrasts in direction and magnitude (Appendix~\ref{app:judge_robustness}), but without repeated generation and judging, paired transitions still describe observed successes and failures rather than stable per-item capabilities, and small differences warrant particular caution. Conversation-level bootstrap intervals quantify benchmark-sample uncertainty only, not generation stochasticity.
\end{document}